\documentclass{article} % For LaTeX2e

\usepackage[final]{neurips_2023}

\usepackage{amsmath,amsfonts,bm}

\def\eqref#1{equation~\ref{#1}}
\def\1{\bm{1}}

\DeclareMathAlphabet{\mathsfit}{\encodingdefault}{\sfdefault}{m}{sl}
\SetMathAlphabet{\mathsfit}{bold}{\encodingdefault}{\sfdefault}{bx}{n}

\usepackage[utf8]{inputenc} % allow utf-8 input
\usepackage[T1]{fontenc}    % use 8-bit T1 fonts
\usepackage{hyperref}       % hyperlinks
\usepackage{url}            % simple URL typesetting
\usepackage{booktabs}       % professional-quality tables
\usepackage{amsfonts}       % blackboard math symbols
\usepackage{nicefrac}       % compact symbols for 1/2, etc.
\usepackage{microtype}      % microtypography
\usepackage[usenames,dvipsnames]{xcolor}         % colors
\hypersetup{
    colorlinks=true,
    linkcolor=black,
    filecolor=black,
    urlcolor=CornflowerBlue,
    citecolor=black,
}
\usepackage{tcolorbox}

\usepackage[colorinlistoftodos,prependcaption,textsize=tiny]{todonotes}

\title{Large Language Models for Automated Data Science:\\
Introducing \methodname{} for Context-Aware Automated Feature Engineering}
\author{
  Noah Hollmann \\
  University of Freiburg \\
  Charit\'e Hospital Berlin \\
  \href{http://priorlabs.ai/}{Prior Labs} \\
  \texttt{noah.homa@gmail.com} \\
  \And
  Samuel M\"uller \\
  \\
  University of Freiburg \\
  \href{http://priorlabs.ai/}{Prior Labs} \\
  \texttt{muellesa@cs.uni-freiburg.de} \\
  \And
  Frank Hutter \\
  \\
  University of Freiburg \\
  \href{http://priorlabs.ai/}{Prior Labs} \\
  \texttt{fh@cs.uni-freiburg.de} \\
}
\usepackage{listings}
\usepackage{xcolor}
\usepackage[utf8]{inputenc}
\usepackage{pgfplots}
\usepackage{float}
\DeclareUnicodeCharacter{2212}{−}
\usepgfplotslibrary{groupplots,dateplot}
\usetikzlibrary{shapes.arrows, positioning, calc, backgrounds, fit}

\pgfplotsset{compat=newest}

\definecolor{codegreen}{rgb}{0,0.6,0}
\definecolor{codegray}{rgb}{0.5,0.5,0.5}
\definecolor{codepurple}{rgb}{0.58,0,0.82}
\definecolor{backcolour}{rgb}{0.95,0.95,0.92}

\usepackage{multirow}

\lstdefinestyle{mystyle}{
    backgroundcolor=\color{white},   
    commentstyle=\color{codegreen},
    keywordstyle=\color{magenta},
    numberstyle=\tiny\color{codegray},
    stringstyle=\color{codepurple},
    basicstyle=\ttfamily\scriptsize,
    breakatwhitespace=false,         
    breaklines=true,                 
    captionpos=b,                    
    keepspaces=true,                 
    numbers=left,                    
    numbersep=5pt,                  
    showspaces=false,                
    showstringspaces=false,
        framextopmargin=0pt, % set left padding
    framexbottommargin=0pt, % set left padding
    showtabs=false,                  
    tabsize=2,
      framexleftmargin=0pt, % set left padding
  framexrightmargin=0pt, % set right padding
    frame=single,
   framerule=0.5pt,
}
\newcommand{\methodname}[0]{CAAFE}

\usepackage{enumitem}
\usepackage{tikz}
\usetikzlibrary{shapes,arrows,positioning}

\usepackage{colortbl}

\begin{document}

\maketitle

\begin{abstract}
As the field of automated machine learning (AutoML) advances, it becomes increasingly important to incorporate domain knowledge into these systems.
We present an approach for doing so by harnessing the power of large language models (LLMs). 
Specifically, we introduce Context-Aware Automated Feature Engineering (\methodname{}), a feature engineering method for tabular datasets that utilizes an LLM to iteratively generate additional semantically meaningful features for tabular datasets based on the description of the dataset. The method produces both Python code for creating new features and explanations for the utility of the generated features.

Despite being methodologically simple, \methodname{} improves performance on 11 out of 14 datasets - boosting mean ROC AUC performance from 0.798 to 0.822 across all dataset - similar to the improvement achieved by using a random forest instead of logistic regression on our datasets.

Furthermore, \methodname{} is interpretable by providing a textual explanation for each generated feature. \methodname{} paves the way for more extensive semi-automation in data science tasks and emphasizes the significance of context-aware solutions that can extend the scope of AutoML systems to semantic AutoML.
We release our %\href{https://github.com/cafeautomatedfeatures/CAFE}{code}
\href{https://github.com/automl/CAAFE}{code}, %\href{https://colab.research.google.com/drive/1JPgiEAKO1e7aofm2vqXA1yTHRhAqZgTx}{a simple demo}
\href{https://colab.research.google.com/drive/1mCA8xOAJZ4MaB_alZvyARTMjhl6RZf0a}{a simple demo}
and a \href{https://pypi.org/project/caafe/}{python package}.
\end{abstract}
% Large Language Models, AutoML, Data Science Automation, Feature Engineering, Domain Knowledge Integration, Tabular Datasets, Explainability

\begin{figure}[h]
%\documentclass{standalone}
%\usepackage{tikz}
%\usetikzlibrary{shapes.geometric, arrows}

%\begin{document}
\centering

\begin{tikzpicture}[block/.style={rectangle, draw, text width=7em, text centered, rounded corners, minimum height=8em},line/.style={draw, -latex},rectconnect/.style={to path={-- ++(0,#1) -| (\tikztotarget)}, rounded corners},]
% Nodes
\node[block, fill=white] (step1) {\textbf{User:} Specifies problem context and dataset};
\node[block, right=0.8cm of step1, fill=white] (step2) {\textbf{LLM:} Generates Code for feature engineering};
\node[block, right=of step2, fill=white] (step3) {\textbf{Interpreter:} Executes generated code};
\node[block, right=of step3, fill=white] (step4) {\textbf{Tabular Prediction Model:}\\Performs cross-validation.};

% Lines
\path[line] (step1) -- (step2);
\path[line] (step2) -- (step3);
\path[line] (step3) -- (step4);
\draw[line, rectconnect=1.5cm] (step4) -- ++(0,2.0cm) -| (step2);

% Evaluation text
\node[align=center, anchor=south] at ($(step4)!0.5!(step2)+(0,1.5cm)$) {Evaluate Performance.\\Keep change if performance is improved.};

% Gray box
\begin{scope}[on background layer]
  \coordinate (top_right) at ($(step4.north east)+(0.25,1.3cm)$);
  \coordinate (bottom_left) at ($(step2.south west)-(0.25cm,0.25cm)$);
  \draw[fill=CornflowerBlue!10] (bottom_left) rectangle (top_right);
  \node[anchor=north east, xshift=-0.1cm, yshift=-0.1cm] at (top_right) {\textbf{CAAFE}};
  %\node[rectangle, draw, inner sep=2pt, anchor=north east] at (top_right) {CAAFE};
\end{scope}
\end{tikzpicture}

%\end{document}
\caption{\methodname{} accepts a dataset as well as user-specified context information and operates by iteratively proposing and evaluating feature engineering operations.}
%\caption{The steps of \methodname{}: The initial steps (1) and (2) are based on context and domain knowledge while the actual predictions are provided by more classical ML models. \methodname{} works in an iterative fashion, accepting code if it passes the evaluation criteria on a validation dataset. Rejected code is provided as a context for further optimization.}
\label{fig:process}
\end{figure}

\section{Introduction}

Automated machine learning (AutoML; e.g., \cite{hutter-book19a}) is very effective at optimizing the machine learning (ML) part of the data science workflow, but existing systems leave tasks such as data engineering and integration of domain knowledge largely to human practitioners.
However, model selection, training, and scoring only account for a small percentage of the time spent by data scientists (roughly 23\% according to the ``State of Data Science''\citep{anaconda2020state}). Thus, the most time-consuming tasks, namely data engineering and data cleaning, are only supported to a very limited degree by AutoML tools, if at all. 
%Crucially, in all automated tasks the integration of real-world domain knowledge is left to the user. 

%AutoML aims to cover the entire data science workflow, from raw dataset processing to deploying a fully-functional machine learning model. Despite its promise, AutoML currently supports only a small portion of the data science workflow, leaving tasks such as data engineering and integration of domain knowledge largely to human practitioners. According to the “State of Data Science” published by Anaconda in 2020, data preprocessing and model selection cover only roughly 11\% of the time spent by a data scientist \citep{anaconda2020state}. However, the most time-consuming tasks, namely data engineering and data cleaning, are only supported to a very limited degree by AutoML tools, if at all. Crucially, in all automated tasks the integration of real-world domain knowledge is left to the user. This approach has been fruitful and appropriate given the technical capabilities of ML tools at the time.

While the traditional AutoML approach has been fruitful and appropriate given the technical capabilities of ML tools at the time, large language models (LLMs) may extend the reach of AutoML to cover more of data science and allow it to evolve towards \emph{automated data science}~\citep{de2022automating}. LLMs encapsulate extensive domain knowledge that can be used to automate various data science tasks, including those that require contextual information. They are, however, not interpretable, or verifiable, and behave less consistently than classical ML algorithms. E.g., even the best LLMs still fail to count or perform simple calculations that are easily solved by classical methods~\citep{gpt-maths, OpenAIHaiku}. 

In this work, we propose an approach that combines the scalability and robustness of classical ML classifiers (e.g. random forests \citep{breiman2001random}) with the vast domain knowledge embedded in LLMs, as visualized in Figure \ref{fig:intro_diagram}. We bridge the gap between LLMs and classical algorithms by using code as an interface between them: LLMs generate code that modifies input datasets, these modified datasets can then be processed by classical algorithms.
Our proposed method, \methodname{}, generates Python code that creates semantically meaningful features that improve the performance of downstream prediction tasks in an iterative fashion and with algorithmic feedback as shown in Figure \ref{fig:process}. Furthermore, CAAFE generates a comment for each feature which explains the utility of generated feature. This allows interpretable AutoML, making it easier for the user to understand a solution, but also to modify and improve on it. Our approach combines the advantages of classical ML (robustness, predictability and a level of interpretability) and LLMs (domain-knowledge and creativity).

% TODO: Not 100% happy with this part yet. I wasn't thinking from automl perspective here
Automating the integration of domain-knowledge into the AutoML process has clear advantages that extend the scope of existing AutoML methods.
These benefits include: 
i) Reducing the latency from data to trained models; ii) Reducing the cost of creating ML models; iii) 
Evaluating a more informed space of solutions than previously possible with AutoML, but a larger space than previously possible with manual approaches for integrating domain knowledge; and
iv) Enhancing the robustness and reproducibility of solutions, as computer-generated solutions are more easily reproduced. \methodname{} demonstrates the potential of LLMs for automating a broader range of data science tasks and highlights the emerging potential for creating more robust and context-aware AutoML tools.

%\todo[inline]{Frank: why do we not use the original diagram with the four quadrants of De Bie et al, which clearly shows that some parts (``data exploration'' and ``exploitation'') are, quote ``more dependent on domain context''? The current figure is suboptimal, since we say LLMs are good for exploratory tasks, but the figure does not put them under data exploration. Ahh maybe you fixed this already!}
%\todo[inline]{Noah: I did not fully understand their graphic, I think data engineering is dependent on context as well, but for them its listed as not really being dependent. Dont you think this is more of a gradient ordered as in the graphic? I do not think we say that LLMs are good for exploratory tasks anywhere? They are good at the more context dependent tasks, but not as heavily context dependent as exploration? Maybe well talk quickly on video?}

\section{Background}
%In this work, we investigate the use of LLMs for feature engineering. To provide context, we outline LLMs and their applications to tabular data tasks, as well as previous feature engineering approaches.

\subsection{Large Language Models (LLMs)}
\label{sec:background:llms}
LLMs are neural networks that are pre-trained on large quantities of raw text data to predict the next word in text documents.
Recently, GPT-4 has been released as a powerful and publicly available LLM \citep{openai2023gpt4}. The architecture of GPT-4 is not oublished, it is likely based on a deep neural network that uses a transformer architecture \citep{VaswaniSPUJGKP17}, large-scale pre-training on a diverse corpus of text and fine-tuning using reinforcement learning from human feedback (RLHF) \citep{ziegler2019fine}. It achieves state-of-the-art performance on various tasks, such as text generation, summarization, question answering and coding. One can adapt LLMs to a specific task without retraining by writing a prompt \citep{brown2020language, wei2021finetuned}; the model parameters are frozen and the model performs in-context inference tasks based on a textual input that formulates the task and potentially contains examples.

\paragraph{LLMs as Tabular Prediction Models}
\citet{hegselmann2023tabllm} recently showed how to use LLMs for tabular data prediction by applying them to a textual representation of these datasets. A prediction on an unseen sample then involves continuing the textual description of that sample on the target column. However, this method requires encoding the entire training dataset as a string and processing it using a transformer-based architecture, where the computational cost increases quadratically with respect to $N \cdot M$, where $N$ denotes the number of samples and $M$ the number of features. Furthermore, the predictions generated by LLMs are not easily interpretable, and there is no assurance that the LLMs will produce consistent predictions, as these predictions depend directly on the complex and heterogeneous data used to train the models. So far, \citet{hegselmann2023tabllm} found that their method yielded the best performance on tiny datasets with up to 8 samples, but was outperformed for larger data sets. 

\paragraph{LLMs for Data Wrangling}
\citet{narayan2022foundation} demonstrated state-of-the-art results using LLMs for entity matching, error detection, and data imputation using prompting and manually tuning the LLMs. \citet{vos2022towards} extended this technique by employing an improved prefix tuning technique.
Both approaches generate and utilize the LLMs output for each individual data sample, executing a prompt for each row. This is in contrast to \methodname{}, which uses code as an interface, making our work much more scalable and faster to execute, since one LLM query can be applied to all samples.

\begin{figure}[t]
% if we stick with this figure, we should fix the alginment here (the descriptions are off to the right a bit right now)
%\documentclass[]{article}
%\usepackage{tikz}
%\usepackage{pgfplots}
%\usetikzlibrary{shadings,backgrounds}
%\usepgfplotslibrary{groupplots,dateplot}
%\usetikzlibrary{shapes.arrows, positioning, calc, backgrounds, fit}

\definecolor{caafecolor}{RGB}{0,0,0}%{44,102,249}
\definecolor{automlcolor}{RGB}{0,0,0}%{59,209,27}

%\begin{document}

\centering
\begin{tikzpicture}[
box/.style={
draw,
minimum width=3.25cm,
minimum height=3.1cm,
text width=2.9cm,
align=center,
font=\small,
rounded corners=5pt,
text depth=1.5cm
},
arrow/.style={
-latex,
line width=1pt
},
scalelabel/.style={
font=\small
}
]
\hspace{-0.6cm}
\node[box, align=center] (dataexploration) {\textbf{Context specification}\\ contextual information, goal analysis and specification};
\node[box, right=0.3cm of dataexploration] (exploitation) {\textbf{Exploitation}\\ interpretation, visualization, reporting, predictions, monitoring};
\node[box, right=0.3cm of exploitation] (dataengineering) {\textbf{Data Engineering}\\ wrangling, integration, preparation, transformation};
\node[box, right=0.3cm of dataengineering] (modelbuilding) {\textbf{Model Building}\\ algorithm choice, parameter tuning, evaluation, model selection};

% Add boxes around the leftmost boxes
%\node[draw, dashed, fit=(dataengineering), inner sep=5pt, label=above:{CAAFE}] {};
\draw[draw=caafecolor, line width=1pt, dotted] ([yshift=20pt, xshift=-2pt]dataengineering.north west) rectangle ([yshift=-2pt, xshift=2pt]dataengineering.south east);
\node[above=3pt, text=caafecolor] at (dataengineering.north) {\small CAAFE};

\draw[draw=automlcolor, line width=1pt, dotted] ([yshift=20pt, xshift=-2pt]modelbuilding.north west) rectangle ([yshift=-2pt, xshift=2pt]modelbuilding.south east);
\node[above=3pt, text=automlcolor] at (modelbuilding.north) {\small Traditional AutoML};

% if we stick with this figure, we should fix the alginment here (the descriptions are off to the right a bit right now)
\node[scalelabel, below left=0.6cm and -1.4cm of dataexploration.south] (classical) {\parbox[c][0.8cm][c]{3cm}{\centering User driven}};
\node[scalelabel, below right=0.6cm and -8.5cm of modelbuilding.south] (fuzzy) {\parbox[c][0.8cm][c]{5cm}{\centering Strength of LLMs: \\ Exploratory, context-dependent, \\ real-world knowledge}};
\node[scalelabel, below right=0.6cm and -2.7cm of modelbuilding.south] (user) {\parbox[c][0.8cm][c]{4cm}{\centering Strength of classical algorithms: \\ Well-defined, predictable}};

%\shade[shading=axis, left color=green, right color=blue, shading angle=90] (dataexploration.west |-fuzzy.north) rectangle (modelbuilding.east|-fuzzy.south);

%\draw[<->, arrow] (classical.east) -- (fuzzy.west);
%\draw[<->, arrow] (fuzzy.east) -- (user.west);

\end{tikzpicture}
%\end{document}
\caption{Data Science pipeline, inspired by \citet{de2022automating}. CAAFE allows for automation of semantic data engineering, while LLMs could provide even further automation: (1) Context specification is user driven (2) exploitation and data engineering can be automated through LLMs (3) model building can be automated by classical AutoML approaches.}
\label{fig:intro_diagram}
\end{figure}

\subsection{Feature Engineering} 
Feature engineering refers to the process of constructing suitable features from raw input data, which can lead to improved predictive performance. 
Given a dataset $D = {(x_i, y_i)}_{i=1}^n$, the goal is to find a function $\phi: \mathcal{X} \rightarrow \mathcal{X'}$ which maximizes the performance of $A(\phi(x_i), y_i)$ for some learning algorithm $A$.
Common methods include numerical transformations, categorical encoding, clustering, group aggregation, and dimensionality reduction techniques, such as principal component analysis \citep{wold1987principal}.

Deep learning methods are capable of learning suitable transformations from the raw input data making them more data-driven and making explicit feature engineering less critical, but only given a lot of data.
%For many tabular setups tree-based approaches are still state-of-the-art \citep{mcelfresh2023neural, grinsztajn2022treebased}.
Thus, appropriate feature engineering still improves the performance of classical and deep learning models, particularly for limited data, complex patterns, or model interpretability.%Furthermore, for tabular data, these classical deep learning methods are still outperformed, which can in part be attributed to the rotation invariance (i.e. implicit feature combinations) of neural networks \citep{grinsztajn2022treebased} % , according to the "No free lunch theorem" \citep{wolpert1997no}

Various strategies for automated feature engineering have been explored in prior studies. Deep Feature Synthesis (DFS; \citet{kanter2015deep}) integrates multiple tables for feature engineering by enumerating potential transformations on features and performing feature selection based on model performance. Cognito \citep{khurana2016cognito} proposes a tree-like exploration of the feature space using handcrafted heuristic traversal strategies.
AutoFeat \citep{horn2019autofeat} employs an iterative subsampling of features using beam search.
Learning-based methods, such as LFE \citep{nargesian2017learning}, utilize machine learning models to recommend beneficial transformations while other methods use reinforcement learning-based strategies \citep{khurana2018feature, zhang2019automatic}. %, in part akin to approaches used in neural architecture search~\citep{chen2019neural}. 
Despite these advancements, none of the existing methods can harness semantic information in an automated manner. %Since the possible space of features to evaluate is huge we propose to use this semantic information to reduce computational efficiency and mitigate multiple-testing issues in feature selection.

\subsubsection{Incorporating Semantic Information}
\begin{figure}[h]
\centering
\pgfplotsset{compat=1.17}

\begin{tikzpicture}
\begin{axis}[
    title={Initial data},
    ylabel={},
    ymin=-1,
    ymax=1,
    xmin=1,
    xmax=14,
    xtick={0,1,2,3,4,5,6,7,8,9,10,11,12,13,14},
    ytick={}, % Remove y-axis ticks by setting it to an empty set
    yticklabels={},
    legend style={at={(0.5,-0.2)},anchor=north},
    legend columns=-1,
    width=6.5cm,
    height=4cm,
]
\addplot[only marks, mark=*, blue] coordinates {
(2,0) (3,0) (4,0) (5,0) (6,0)  (9,0) (10,0) (11,0) (12,0) (13,0)};
\addplot[only marks, mark=*, orange] coordinates {(1,0) (7,0) (14,0)};
\addplot[only marks, mark=*, green, fill=white] coordinates {(8,0)};
\legend{True, False, Test-Sample};

\end{axis}
\end{tikzpicture}
\hspace{1cm}
\begin{tikzpicture}
\begin{axis}[
    title={With extra feature based on context},
    ylabel={},%{Weekday (1) or Weekend (0)},
    ylabel style={text width=2.5cm},
    ymin=-0.5,
    ymax=1.5,
    xmin=1,
    xmax=14,
    xtick={0,1,2,3,4,5,6,7,8,9,10,11,12,13,14},
    ytick={0,1},
    yticklabels={Weekday, Weekend},
    legend style={at={(0.5,-0.2)},anchor=north},
    legend columns=-1,
    width=6.5cm,
    height=4cm,
]
\addplot[only marks, mark=*, blue] coordinates {
(2,0) (3,0) (4,0) (5,0) (6,0) (9,0) (10,0) (11,0) (12,0) (13,0)};
\addplot[only marks, mark=*, orange] coordinates {(1,1) (7,1) (14,1) 
};
\addplot[only marks, mark=o, green] coordinates {(8,1)};
\legend{Working, Day-Off, Test-Sample};

\end{axis}
\end{tikzpicture}
\caption{%Usefulness of contextual information for feature engineering:
Contextual information can simplify a task immensely.
On the left-hand side no contextual information is added to the plot, and it is hard to predict the label for the green query point.
On the right-hand side contextual information is added and a useful additional feature (weekend or weekday) is derived from which a mapping from features to targets can be found.}
\label{fig:feature_extension_demo}
\end{figure}

The potential feature space, when considering the combinatorial number of transformations and combinations, is vast. Therefore, semantic information is useful, to serve as a prior for identifying useful features. By incorporating semantic and contextual information, feature engineering techniques can be limited to semantically meaningful features enhancing the performance by mitigating issues with multiple testing and computational complexity and boosting the interpretability of machine learning models. This strategy is naturally applied by human experts who leverage their domain-specific knowledge and insights.
Figure \ref{fig:feature_extension_demo} exemplifies the usefulness of contextual information.

\lstset{style=mystyle}
\lstdefinestyle{mystyle2}{
    backgroundcolor=\color{white},   
    %commentstyle=\color{codegreen},
    %keywordstyle=\color{magenta},
    %numberstyle=\tiny\color{codegray},
    %stringstyle=\color{codepurple},
    basicstyle=\ttfamily\scriptsize\color{black},
    breakatwhitespace=false,         
    breaklines=true,                 
    captionpos=b,                    
    keepspaces=true,                     
    showspaces=false,                
    showstringspaces=false,
    framextopmargin=3pt, % set left padding
    framexbottommargin=3pt, % set left padding
  framexleftmargin=3pt, % set left padding
  framexrightmargin=3pt, % set right padding
    numbers=none,
    showtabs=false,                  
    tabsize=2,
    frame=single,
   framerule=1pt,
}
\lstset{style=mystyle2}
\definecolor{lightblue}{RGB}{200,200,255}
\definecolor{lightred}{RGB}{255,200,200}
\begin{figure}
\input{figures/example_run}
    \caption{Exemplary run of \methodname{} on the Tic-Tac-Toe Endgame dataset. User generated input is shown in \textcolor{blue}{blue}, ML-classifier generated data shown in \textcolor{red}{red} and LLM generated code is shown with syntax highlighting. The generated code contains a comment per generated feature that follows a template provided in our prompt (Feature name, description of usefulness, features used in the generated code and sample values of these features). In this run, \methodname{} improves the ROC AUC on the validation dataset from 0.888 to 1.0 in two feature engineering iterations.}
    \label{fig:example_run}
\end{figure}

\section{Method}
\label{sec:method}
We present \methodname{}, an approach that leverages large language models to incorporate domain knowledge into the feature engineering process, offering a promising direction for automating data science tasks while maintaining interpretability and performance.

Our method takes the training and validation datasets, $D_{train}$ and $D_{valid}$, as well as a description of the context of the training dataset and features as input. From this information \methodname{} constructs a prompt, i.e. instructions to the LLM containing specifics of the dataset and the feature engineering task.
Our method performs multiple iterations of feature alterations and evaluations on the validation dataset, as outlined in Figure \ref{fig:process}.
%In the $i$th iteration  the LLM generates code $C$ based on our prompt, which is then executed on both $D^{i}_{train}$ and $D^{i}_{valid}$ and the transformed datasets $D^{*}_{train}$ and $D^{*}_{valid}$ are retrieved. The transformed dataset $D^{*}_{train}$ is then used to fit a traditional ML-classifier and predict on $D^{*}_{valid}$, yielding a performance $P^{*}$. If $P^{*}$ exceeds the performance $P^{i}$ for training on $D^{i}_{train}$, the feature is kept and $D^{i}_{train} := D^{*}_{train}$. Otherwise the feature is rejected and $D^{i+1}_{train} := D^{i}_{train}$. Figure \ref{fig:example_run} shows a shortened version of one such run on the Tic-Tac-Toe Endgame dataset. Figure \ref{fig:llm_prompt} in the appendix shows an exemplary full prompt.
In each iteration, the LLM generates code, which is then executed on the current $D_{train}$ and $D_{valid}$ resulting in the transformed datasets $D'_{train}$ and $D'_{valid}$. We then use $D'_{train}$ to fit an ML-classifier and evaluate its performance $P'$ on $D'_{valid}$. If $P'$ exceeds the performance $P$ achieved by training on $D_{train}$ and evaluating on $D_{valid}$, the feature is kept and we set $D_{train} := D'_{train}$ and $D_{valid} := D'_{valid}$. Otherwise, the feature is rejected and $D_{train}$ and $D_{valid}$ remain unchanged. Figure \ref{fig:example_run} shows a shortened version of one such run on the Tic-Tac-Toe Endgame dataset. 

\paragraph{Prompting LLMs for Feature Engineering Code}
\label{ssec:prompting:method}

Here, we describe how \methodname{} builds the prompt that is used to perform feature engineering. In this prompt, the LLM is instructed to create valuable features for a subsequent prediction task and to provide justifications for the added feature's utility. It is also instructed to drop unnecessary features, e.g. when their information is captured by other created features.

The prompt contains semantic and descriptive information about the dataset. Descriptive information, i.e.\ summary statistics, such as the percentage of missing values is based solely on the train split of the dataset.
The prompt consists of the following data points:
\renewcommand{\theenumi}{\Alph{enumi}}
\begin{enumerate}[label=\Alph*, font=\bfseries]
\item A user-generated dataset description, that contains contextual information about the dataset (see Section \ref{sec:experimental_setup} for details on dataset descriptions for our experiments)
\item Feature names adding contextual information and allowing the LLM to generate code to index features by their names
\item Data types (e.g. float, int, category, string) - this adds information on how to handle a feature in the generated code
\item Percentage of missing values - missing values are an additional challenge for code generation%. Missing values are often corner cases in the generated functions that can lead to unexpected behavior
\item 10 random rows from the dataset - this provides information on the feature scale, encoding, etc. %If the dataset contains sensitive information, this should not be included in the prompt when there is transmission of the prompt over the web. Based solely on the train split of the dataset.
\end{enumerate}

Additionally, the prompt provides a template for the expected form of the generated code and explanations. Adding a template when prompting is a common technique to improve the quality of responses \citep{openaiop3:online}. We use Chain-of-thought instructions -- instructing a series of intermediate reasoning steps --, another effective technique for prompting \citep{wei2023chainofthought}. The prompt includes an example of one such Chain-of-thought for the code generation of one feature: first providing the high-level meaning and usefulness of the generated feature, providing the names of features used to generate it, retrieving sample values it would need to accept and finally writing a line of code. We provide the complete prompt in Figure \ref{fig:llm_prompt} in the appendix.
% TODO: Maybe include the comment template here?

If the execution of a code block raises an error, this error is passed to the LLM for the next code generation iteration. We observe that using this technique \methodname{} recovered from all errors in our experiments. One such example can be found in Table \ref{tab:strategies}.

\paragraph{Technical Setup}
\label{ssec:experimental_setup:method}
The data is stored in a Pandas dataframe \citep{mckinney-proc-scipy-2010}, which is preloaded into memory for code execution. The generated Python code is executed in an environment where the training and validation data frame is preloaded.
The performance is measured on the current dataset with ten random validation splits $D_{valid}$ and the respective transformed datasets $D'_{valid}$ with the mean change of accuracy and ROC AUC used to determine if the changes of a code block are kept, i.e.\ when the average of both is greater than 0.
We use OpenAI's GPT-4 and GPT-3.5 as LLMs \citep{openai2023gpt4} in \methodname{}.
%GPT-4 is the latest and most advanced system developed by OpenAI. 
We perform ten feature engineering iterations and TabPFN \citep{hollmann2022tabpfn} in the iterative evaluation of code blocks. 

The automatic execution of AI-generated code carries inherent risks, such as misuse by malicious actors or unintended consequences from AI systems operating outside of controlled environments.
Our approach is informed by previous studies on AI code generation and cybersecurity \citep{AI_Code_Generation_and_Cybersecurity_2023, AI_Attack_Vector_2023}.
We parse the syntax of the generated python code and use a whitelist of operations that are allowed for execution. Thus operations such as imports, arbitrary function calls and others are excluded. This does not provide full security, however, e.g. does not exclude operations that can lead to infinite loops and excessive resource usage such as loops and list comprehensions.

\section{Experimental Setup}
\label{sec:experimental_setup}

%In this section, we outline the experimental setup used to evaluate the performance of our proposed method, \methodname{}. We discuss the datasets, preprocessing techniques, and evaluation metrics employed in our experiments.

\paragraph{Setup of Downstream-Classifiers}
%\todo[inline]{Frank: For the NeurIPS submission, it would indeed be nice to evaluate the effect on more classifiers here, including XGBoost, AutoGluon and Auto-sklearn2.}
We evaluate our method with Logistic Regression, Random Forests \citep{breiman2001random} and TabPFN \citep{hollmann2022tabpfn} for the final evaluation while using TabPFN to evaluate the performance of added features. We impute missing values with the mean, one-hot or ordinal encoded categorical inputs, normalized
features and passed categorical feature indicators, where necessary, using the setup of \citet{hollmann2022tabpfn} \footnote{\url{https://github.com/automl/TabPFN/blob/main/tabpfn/scripts/tabular_baselines.py}}.

\paragraph{Setup of Automated Feature Engineering Methods}

%We do not claim to outperform these baselines but to extend the capabilities of downstream classifiers and to solve a distinctly different task to previous automated feature engineering methods.
%Even when no fair comparison can be made, 
We also evaluate popular context-agnostic feature engineering libraries Deep Feature Synthesis (DFS; \citealp{kanter2015deep}) and AutoFeat \citep{horn2019autofeat}\footnote{\url{https://github.com/alteryx/featuretools}, \url{https://github.com/cod3licious/autofeat}}. We evaluate DFS and AutoFeat alone and in combination with \methodname{}. When combined, \methodname{} is applied first and the context-agnostic AutoFE method subsequently. For DFS we use the primitives "add\_numeric" and "multiply\_numeric", and default settings otherwise. For TabPFN, DFS generates more features than TabPFN accepts (the maximum number of features is 100) in some cases.
%would be nice to have a number here, as otherwise one could say that probably DFS does not help for Tabpfn just because it creates too many features very often
In these cases, we set the performance to the performance without feature engineering. For AutoFeat, we use one feature engineering step and default settings otherwise.

\paragraph{Evaluating LLMs on Tabular Data}
The LLM's training data originates from the web, potentially including datasets and related notebooks. GPT-4 and GPT-3.5 have a knowledge cutoff in September 2021, i.e., almost all of its training data originated from before this date. Thus, an evaluation on established benchmarks can be biased since a textual description of these benchmarks might have been used in the training of the LLM.

We use two categories of datasets for our evaluation: (1) widely recognized datasets from OpenML released before September 2021, that could potentially be part of the LLMs training corpus and (2) lesser known datasets from Kaggle released after September 2021 and only accessible after accepting an agreement and thus harder to access by web crawlers.

From OpenML~\citep{OpenML2013,OpenMLPython2019}, we use small datasets that have descriptive feature names (i.e. we do not include any datasets with numbered feature names). Datasets on OpenML contain a task description that we provide as user context to our method.
When datasets are perfectly solvable with TabPFN alone (i.e. reaches ROC AUC of 1.0) we reduce the training set size for that dataset, marked in Table \ref{table:results_table_per_dataset}.
We focus on small datasets with up to $2\,000$ samples in total, because feature engineering is most important and significant for smaller datasets.

We describe the collection and preprocessing of datasets in detail in Appendix \ref{sec:datasets}. %We believe that an evaluation of the common datasets in (1) is still informative since the feature engineering and code generation task is an abstraction from simply seeing the datasets which would likely be less affected by having seen the data before.

% TODO: Add data preprocessing

% Frank: we should only cite this if we use a particular benchmarking suite: bischl-neuripsdbt21a
% \todo[inline]{Frank: we should state exactly how we chose whether to include a dataset or not. The constraints you mention below are great, but which set of datasets did you start with? The joint valid+test set from the TabPFN paper?}

\paragraph{Evaluation Protocol}
% TODO: Extend
\label{ssec:experimental_setup:protocol}
For each dataset, we evaluate $5$ repetitions, each with a different random seed and train- and test split to reduce the variance stemming from these splits~\citep{MLSYS2021_cfecdb27}. We split into 50\% train and 50\% test samples and all methods used the same splits.

\begin{table}
    \centering
    \small
    \caption{ROC AUC OVO results using TabPFN. $\pm$ indicates the standard deviation across 5 splits. [R] indicates datasets where reduced data was used because TabPFN had 100\% accuracy by default, see Appendix \ref{sec:datasets}.}
    \label{table:results_table_per_dataset}
    \begin{tabular}{l|r|r|r}
\toprule
{} & \multicolumn{3}{c}{TabPFN} \\
{} & {No Feat. Eng.} & CAAFE (GPT-3.5) & CAAFE (GPT-4) \\
\midrule
airlines                   &  \textbf{0.6211} {\scriptsize $\pm$.04} &            0.619 {\scriptsize $\pm$.04} &           0.6203 {\scriptsize $\pm$.04} \\
balance-scale [R]              &           0.8444 {\scriptsize $\pm$.29} &            0.844 {\scriptsize $\pm$.31} &   \textbf{0.882} {\scriptsize $\pm$.26} \\
breast-w [R]                   &           0.9783 {\scriptsize $\pm$.02} &  \textbf{0.9809} {\scriptsize $\pm$.02} &  \textbf{0.9809} {\scriptsize $\pm$.02} \\
cmc                        &           0.7375 {\scriptsize $\pm$.02} &           0.7383 {\scriptsize $\pm$.02} &  \textbf{0.7393} {\scriptsize $\pm$.02} \\
credit-g                   &           0.7824 {\scriptsize $\pm$.03} &           0.7824 {\scriptsize $\pm$.03} &  \textbf{0.7832} {\scriptsize $\pm$.03} \\
diabetes                   &           0.8427 {\scriptsize $\pm$.03} &  \textbf{0.8434} {\scriptsize $\pm$.03} &           0.8425 {\scriptsize $\pm$.03} \\
eucalyptus                 &  \textbf{0.9319} {\scriptsize $\pm$.01} &           0.9317 {\scriptsize $\pm$.01} &  \textbf{0.9319} {\scriptsize $\pm$.00} \\
jungle\_chess..             &           0.9334 {\scriptsize $\pm$.01} &           0.9361 {\scriptsize $\pm$.01} &  \textbf{0.9453} {\scriptsize $\pm$.01} \\
pc1                        &           0.9035 {\scriptsize $\pm$.01} &           0.9087 {\scriptsize $\pm$.02} &  \textbf{0.9093} {\scriptsize $\pm$.01} \\
tic-tac-toe [R]                &           0.6989 {\scriptsize $\pm$.08} &           0.6989 {\scriptsize $\pm$.08} &  \textbf{0.9536} {\scriptsize $\pm$.06} \\
$\langle Kaggle\rangle$ health-insurance  &           0.5745 {\scriptsize $\pm$.02} &           0.5745 {\scriptsize $\pm$.02} &  \textbf{0.5748} {\scriptsize $\pm$.02} \\
$\langle Kaggle\rangle$ pharyngitis       &           0.6976 {\scriptsize $\pm$.03} &           0.6976 {\scriptsize $\pm$.03} &  \textbf{0.7078} {\scriptsize $\pm$.04} \\
$\langle Kaggle\rangle$ kidney-stone      &           0.7883 {\scriptsize $\pm$.04} &           0.7873 {\scriptsize $\pm$.04} &  \textbf{0.7903} {\scriptsize $\pm$.04} \\
$\langle Kaggle\rangle$ spaceship-titanic &            0.838 {\scriptsize $\pm$.02} &           0.8383 {\scriptsize $\pm$.02} &  \textbf{0.8405} {\scriptsize $\pm$.02} \\
%\midrule
%Mean ROC AUC                  &            0.798 {\scriptsize $\pm$.05} &           0.7987 {\scriptsize $\pm$.05} &  \textbf{0.8215} {\scriptsize $\pm$.04} \\
%Mean ROC AUC Rank                  &                                      2.43 &                                      2.32 &                             \textbf{1.25} \\
\bottomrule
\end{tabular}

\end{table}

\section{Results}
% TODO: Needs more work
\label{sec:results}
%In this section, we present an empirical analysis of our method on real-world classification tasks. We demonstrate that large gains in predictor performance can be made using \methodname{}.

%\begin{figure}[t]
%\include{figures/results_table}
%\caption{XYZ}
%\label{tab:results_table}
%\end{figure}
% [Table per dataset: Number of feats extended, Diff in performance, kind of dataset and description]

%\subsubsection{Observed Feature Engineering Strategies}
In this section we showcase the results of our method in three different ways.
First, we show that \methodname{} can improve the performance of a state-of-the-art classifier. Next, we show how \methodname{} interacts with traditional automatic feature engineering methods and conclude with examples of the features that \methodname{} creates.

\begin{table}[H]
    \centering
    \caption{Mean ROC AUC and average rank (ROC AUC) per downstream classification method and feature extension method. Best AutoFE method per base classifer is shown in bold. The features generated by \methodname{} are chosen with TabPFN as classifier. Rank are calculated across all classifiers and feature engineering methods. FETCH was too computationally expensive to compute for all base classifiers in the rebuttal. Each seed and dataset takes up to 24 hours and has to be evaluated for each base classifer independently. Thus, we use features computed for logistic regression for all other classifiers.}
    \scriptsize
    \begin{table}[H]
    \label{table:proposal}
    \centering
    \scriptsize
    \setlength{\tabcolsep}{5pt}
\begin{tabular}{ll|l|llll|ll}
\toprule
{} & {} & {} & \multicolumn{4}{c}{Baselines} & \multicolumn{2}{c}{CAAFE} \\
{} & {} & {No FE} & DFS & AutoFeat & FETCH & OpenFE & GPT-3.5 & GPT-4 \\

\midrule
Log. Reg. & Mean &  0.749 &          0.764 &    0.754 &   0.76 &  0.757 &          0.763 &  \textbf{0.769} \\
\rowcolor{gray!25}            & Mean Rank &   27.4 &  \textbf{23.6} &     26.2 &   25.2 &     25 &           24.8 &            24.3 \\
\midrule
Random Forest & Mean &  0.782 &          0.783 &    0.783 &  0.785 &  0.785 &           0.79 &  \textbf{0.803} \\
\rowcolor{gray!25}            & Mean Rank &   23.4 &           22.1 &     21.8 &   23.5 &   22.3 &           23.1 &   \textbf{19.9} \\
\midrule
ASKL2 & Mean &  0.807 &          0.801 &    0.808 &  0.807 &  0.806 &          0.815 &  \textbf{0.818} \\
\rowcolor{gray!25}            & Mean Rank &   12.2 &           12.9 &     12.6 &   13.4 &   13.5 &  \textbf{10.9} &            11.6 \\
\midrule
Autogluon & Mean &  0.796 &          0.799 &    0.797 &  0.787 &  0.798 &          0.803 &  \textbf{0.812} \\
\rowcolor{gray!25}            & Mean Rank &   17.6 &           15.4 &     16.4 &   17.6 &   16.6 &           15.8 &   \textbf{14.1} \\
\midrule
TabPFN & Mean &  0.798 &          0.791 &    0.796 &  0.796 &  0.798 &          0.806 &  \textbf{0.822} \\
\rowcolor{gray!25}            & Mean Rank &   13.9 &             15 &     14.8 &   16.5 &   13.9 &           12.9 &   \textbf{9.78} \\
\bottomrule
\end{tabular}
\end{table}
    \label{table:results_table_means}
\end{table}

\iffalse
\begin{table}[b]
    \centering
    \caption{Mean ROC AUC and average rank (ROC AUC) per downstream classification method and feature extension method. The features generated by \methodname{} are chosen with TabPFN as classifier. $\pm$ indicates the standard deviation across 5 splits. AutoFeat and DFS improve simpler classifiers but not TabPFN, while \methodname{} improves all setups. %For the average ranks, we refer to Appendix \ref{sec:ranks_methods_and_prompts}.
    \label{table:results_table_means}
    }
    \scriptsize
    \input{figures/results_cafe_and_baselines}
\end{table}
\fi
\begin{table}
    \centering
    \caption{Examples of common strategies employed by \methodname{} for feature extension. The full code and comments are automatically generated based on the user-provided dataset descriptions. \methodname{} combines features, creates ordinal versions of numerical features through binning, performs string transformations, removes superfluous features, and even recovers from errors when generating invalid code.}
    \label{tab:strategies}
    \begin{tabular}{lll}
        \toprule
        Description & Generated code \\
        \midrule
        \addlinespace
        \begin{minipage}[t]{0.25\textwidth}
         \textbf{Combination} \\ \scriptsize Example from the Kaggle Kidney Stone dataset.
        \end{minipage}
        &
        \begin{minipage}[t]{0.72\textwidth}
        \vspace{-4mm}
        \begin{lstlisting}[language=Python, basicstyle=\scriptsize, frame=single, rulecolor=\color{black}]
# Usefulness: Fever and rhinorrhea are two of the most common symptoms of respiratory infections, including GAS pharyngitis. This feature captures their co-occurrence.
# Input samples: 'temperature': [38.0, 39.0, 39.5], 'rhinorrhea': [0.0, 0.0, 0.0]
df['fever_and_rhinorrhea'] = ((df['temperature'] >= 38.0) & (df['rhinorrhea'] > 0)).astype(int)\end{lstlisting}
        \vspace{-4mm}
        \end{minipage} \\
        \addlinespace
        \addlinespace
        \addlinespace
        \begin{minipage}[t]{0.25\textwidth}
        \textbf{Binning} \\ \scriptsize Example from the Kaggle Spaceship Titanic dataset.
        \end{minipage}
        &
        \begin{minipage}[t]{0.72\textwidth}
        \vspace{-4mm}
        \begin{lstlisting}[language=Python, basicstyle=\scriptsize, frame=single, rulecolor=\color{black}]
# Feature: AgeGroup (categorizes passengers into age groups)
# Usefulness: Different age groups might have different likelihoods of being transported.
# Input samples: 'Age': [30.0, 0.0, 37.0]
bins = [0, 12, 18, 35, 60, 100]
labels = ['Child', 'Teen', 'YoungAdult', 'Adult', 'Senior']
df['AgeGroup'] = pd.cut(df['Age'], bins=bins, labels=labels)
df['AgeGroup'] = df['AgeGroup'].astype('category')\end{lstlisting}
        \vspace{-4mm}
        \end{minipage} \\
        \addlinespace
        \addlinespace
        \addlinespace
        \begin{minipage}[t]{0.25\textwidth}
        \textbf{String transformation} \\ \scriptsize Example from the Kaggle Spaceship Titanic dataset.
        %The original "Cabin" field is hard to use with a standard classifier, as this is the room number which is almost unique for each person. \methodname{} splits up the more interpretable deck and side features, which actually then do appear more than once in the data.
        \end{minipage}
        &
        \begin{minipage}[t]{0.72\textwidth}
        \vspace{-4mm}
        \begin{lstlisting}[language=Python, basicstyle=\scriptsize, frame=single, rulecolor=\color{black}]
# Feature: Deck
# Usefulness: The deck information can help identify patterns in the location of cabins associated with transported passengers.
# Input samples: 'Cabin': ['F/356/S', 'G/86/P', 'C/37/P']
df['Deck'] = df['Cabin'].apply(lambda x: x[0] if isinstance(x, str) else 'Unknown')

# Feature: CabinSide
# Usefulness: The side of the cabin can help identify patterns in the location of cabins associated with transported passengers.
# Input samples: 'Cabin': ['F/356/S', 'G/86/P', 'C/37/P']
df['CabinSide'] = df['Cabin'].apply(lambda x: x.split('/')[-1] if isinstance(x, str) else 'Unknown')\end{lstlisting}
        \vspace{-4mm}
        \end{minipage} \\
        \addlinespace
        \addlinespace
        \addlinespace
        \begin{minipage}[t]{0.25\textwidth}
        \textbf{Removing features} \\ \scriptsize Example from the Balance Scale dataset.
        \end{minipage}
        &
        \begin{minipage}[t]{0.72\textwidth}
        \vspace{-4mm}
        \begin{lstlisting}[language=Python, basicstyle=\scriptsize, frame=single, rulecolor=\color{black}]
# Drop original columns
# Explanation: The original columns 'left-weight', 'left-distance', 'right-weight', and 'right-distance' can be dropped as the newly created columns 'left_moment', 'right_moment', and 'moment_difference' capture the relevant information for predicting the 'class'.
df.drop(columns=['left-weight', 'left-distance', 'right-weight', 'right-distance'], inplace=True)
# No more additional columns needed\end{lstlisting}
        \end{minipage} \\
        \addlinespace
        \begin{minipage}[t]{0.25\textwidth}
        \textbf{Error Recovery} \\ \scriptsize Example from the Kaggle Health Insurance dataset. The code execution failed because missing values were not considered in the dataframe. In the next iteration code is suggested that can handle missing values.
        \end{minipage}
        &
        \begin{minipage}[t]{0.72\textwidth}
        \vspace{-4mm}
        \begin{lstlisting}[language=Python, basicstyle=\scriptsize, frame=single, rulecolor=\color{black}]
# Feature: Age_difference
# Usefulness: Age difference between upper and lower age can be useful in determining the likelihood of a person applying for a policy.
# Input samples: 'Upper_Age': ['C33', 'C2', 'C3'], 'Lower_Age': ['Owned', 'Rented', 'Rented']
df['Age_difference'] = df['Upper_Age'].str.extract('(\d+)').astype(int) - df['Lower_Age'].str.extract('(\d+)').astype(int)

Feedback: failed with error: cannot convert float NaN to integer. 

# Feature: Age_difference
# Usefulness: Age difference between upper and lower age can be useful in determining the likelihood of a person applying for a policy.
# Input samples: 'Upper_Age': ['C33', 'C2', 'C3'], 'Lower_Age': ['Owned', 'Rented', 'Rented']
df['Age_difference'] = df['Upper_Age'].str.extract('(\d+)').astype(float).fillna(0) - df['Lower_Age'].str.extract('(\d+)').astype(float).fillna(0)\end{lstlisting}
        \end{minipage} \\
        \bottomrule
    \end{tabular}
\end{table}

\paragraph{Performance of \methodname{}} %Here we discuss results using TabPFN as a downstream classifier and GPT-4 as the LLM, our strongest setup, and discuss other combinations in the following paragraph.
\methodname{} can improve our strongest classifier, TabPFN, substantially. If it is used with GPT-4, we improve average ROC AUC performance from 0.798 to 0.822, as shown in Table \ref{table:results_table_means}, and enhance the performance for 11/14 datasets.% tie on 1/14 and reduce performance on two datasets. 
On the evaluated datasets, this improvement is similar (71\%) to the average improvement achieved by using a random forest (AUC 0.783) instead of logistic regression (AUC 0.749).
We can see that CAAFE even improves performance for all of the new datasets from Kaggle.
If we use CAAFE with GPT-3.5 only, we can see that it performs clearly worse than with GPT-4, and only improves performance on 6/14 datasets.

There is great variability in the improvement size depending on whether (1) a problem is amenable to feature engineering, i.e. is there a mapping of features that explains the data better and that can be expressed through simple code; and (2) the quality of the dataset description (e.g., the balance-scale dataset contains an accurate description of how the dataset was constructed) %or rule-based, so effective features can be found by applying these rules (e.g. tic-tac-toe or jungle\_chess).
Per dataset performance can be found in Table \ref{table:results_table_per_dataset}.
\methodname{} takes 4:43 minutes to run on each dataset, 90\% of the time is spent on the LLM's code generation and 10\% on the evaluation of the generated features. %Running \methodname{} with 10 iterations costs $0.71\$$ per dataset on average.
In Appendix \ref{sec:results_cost_and_time} we plot the performance, time and cost of \methodname{} across feature engineering iterations, showing the tradeoff between these parameters. For the 14 datasets, 5 splits and 10 \methodname{} iterations, \methodname{} generates 52 faulty features (7.4\%) in the generation stage, from which it recovers (see Figure \ref{tab:strategies}).

\paragraph{Incorporating Classical AutoFE Methods} 
Classical AutoFE methods can readily be combined with our method, one simply runs \methodname{} first and then lets a classical AutoFE method find further feature extensions, as we did in Table \ref{table:results_table_means}. 
For less powerful downstream classifiers, namely Logistic Regression and Random Forests, we observe that applying AutoFE additionally to \methodname{} improves performance further. 
%\methodname{} alone likely does not add enough meaningful features in $10$ iterations and so existing AutoFE improve on this.
The AutoML method TabPFN on the other hand is not improved by applying the evaluated AutoFE methods.
This discrepancy might stem from the larger hypothesis space (complexity) of TabPFN, it can get all necessary information from the data directly. % compared to logistic regression for example, which can only model linear functions.
For all combinations of classifiers and additional AutoFE methods, we can see that \methodname{} improves performance on average.
\section{Conclusion}
Our study presents a novel approach to integrating domain knowledge into the AutoML process through Context-Aware Automated Feature Engineering (\methodname{}). By leveraging the power of large language models, \methodname{} automates feature engineering for tabular datasets, generating semantically meaningful features and explanations of their utility. Our evaluation demonstrates the effectiveness of this approach, which complements existing automated feature engineering and AutoML methods.

This work emphasizes the importance of context-aware solutions in achieving robust outcomes. We expect that LLMs will also be useful for automating other aspects of the data science pipeline, such as data collection, processing, model building, and deployment. As large language models continue to improve, it is expected that the effectiveness of \methodname{} will also increase.

Dataset descriptions play a critical role in our method; however, in our study, they were derived solely from web-crawled text associated with public datasets. If users were to provide more accurate and detailed descriptions, the effectiveness of our approach could be significantly improved.

However, our current approach has some limitations. Handling datasets with a large number of features can lead to very large prompts, which can be challenging for LLMs to process effectively. The testing procedure for adding features is not based on statistical tests, and could be improved using techniques of previous feature engineering works. LLMs, at times, exhibit a phenomenon known as "hallucinations.", where models produce inaccurate or invented information. Within CAAFE, this might result in the generation of features and associated explanations that appear significant and are logically presented, even though they may not be grounded in reality. Such behavior can be problematic, especially when individuals place trust in these systems for essential decision-making or research tasks. Finally, the usage of LLMs in automated data analysis comes with a set of societal and ethical challenges. Please see Section \ref{sec:impacts} for a discussion on the broader impact and ethical considerations.

Future research may explore prompt tuning, fine-tuning language models, and automatically incorporating domain-knowledge into models in other ways. Also, there may lie greater value in the interaction of human users with such automated methods, also termed human-in-the-loop AutoML \citep{lee2020human}, where human and algorithm interact continuously. This would be particularly easy with a setup similar to \methodname{}, as the input and output of the LLM are interpretable and easily modified by experts.

\newpage

\bibliography{iclr2023_conference_tinypaper}

\begin{thebibliography}{36}
\providecommand{\natexlab}[1]{#1}
\providecommand{\url}[1]{\texttt{#1}}
\expandafter\ifx\csname urlstyle\endcsname\relax
  \providecommand{\doi}[1]{doi: #1}\else
  \providecommand{\doi}{doi: \begingroup \urlstyle{rm}\Url}\fi

\bibitem[Anaconda(2020)]{anaconda2020state}
Anaconda.
\newblock The state of data science 2020.
\newblock Website, 2020.
\newblock URL \url{https://www.anaconda.com/state-of-data-science-2020}.

\bibitem[Bouthillier et~al.(2021)Bouthillier, Delaunay, Bronzi, Trofimov,
  Nichyporuk, Szeto, Mohammadi~Sepahvand, Raff, Madan, Voleti, Ebrahimi~Kahou,
  Michalski, Arbel, Pal, Varoquaux, and Vincent]{MLSYS2021_cfecdb27}
Xavier Bouthillier, Pierre Delaunay, Mirko Bronzi, Assya Trofimov, Brennan
  Nichyporuk, Justin Szeto, Nazanin Mohammadi~Sepahvand, Edward Raff, Kanika
  Madan, Vikram Voleti, Samira Ebrahimi~Kahou, Vincent Michalski, Tal Arbel,
  Chris Pal, Gael Varoquaux, and Pascal Vincent.
\newblock Accounting for variance in machine learning benchmarks.
\newblock In A.~Smola, A.~Dimakis, and I.~Stoica (eds.), \emph{Proceedings of
  Machine Learning and Systems}, volume~3, pp.\  747--769, 2021.
\newblock URL
  \url{https://proceedings.mlsys.org/paper_files/paper/2021/file/cfecdb276f634854f3ef915e2e980c31-Paper.pdf}.

\bibitem[Breiman(2001)]{breiman2001random}
Leo Breiman.
\newblock Random forests.
\newblock \emph{Machine learning}, 45:\penalty0 5--32, 2001.

\bibitem[Brown et~al.(2020)Brown, Mann, Ryder, Subbiah, Kaplan, Dhariwal,
  Neelakantan, Shyam, Sastry, Askell, et~al.]{brown2020language}
Tom Brown, Benjamin Mann, Nick Ryder, Melanie Subbiah, Jared~D Kaplan, Prafulla
  Dhariwal, Arvind Neelakantan, Pranav Shyam, Girish Sastry, Amanda Askell,
  et~al.
\newblock Language models are few-shot learners.
\newblock \emph{Advances in neural information processing systems},
  33:\penalty0 1877--1901, 2020.

\bibitem[Crockett(2023)]{AI_Attack_Vector_2023}
Adam Crockett.
\newblock Ai generated code creates a new security attack vector, April 2023.
\newblock URL
  \url{https://dev.to/adam_cyclones/ai-generated-code-creates-a-new-security-attack-vector-39if}.

\bibitem[De~Bie et~al.(2022)De~Bie, De~Raedt, Hern{\'a}ndez-Orallo, Hoos,
  Smyth, and Williams]{de2022automating}
Tijl De~Bie, Luc De~Raedt, Jos{\'e} Hern{\'a}ndez-Orallo, Holger~H Hoos,
  Padhraic Smyth, and Christopher~KI Williams.
\newblock Automating data science.
\newblock \emph{Communications of the ACM}, 65\penalty0 (3):\penalty0 76--87,
  2022.

\bibitem[Feurer et~al.()Feurer, van Rijn, Kadra, Gijsbers, Mallik, Ravi,
  Mueller, Vanschoren, and Hutter]{OpenMLPython2019}
Matthias Feurer, Jan~N. van Rijn, Arlind Kadra, Pieter Gijsbers, Neeratyoy
  Mallik, Sahithya Ravi, Andreas Mueller, Joaquin Vanschoren, and Frank Hutter.
\newblock Openml-python: an extensible python api for openml.
\newblock \emph{arXiv}, 1911.02490.
\newblock URL \url{https://arxiv.org/pdf/1911.02490.pdf}.

\bibitem[Hegselmann et~al.(2023)Hegselmann, Buendia, Lang, Agrawal, Jiang, and
  Sontag]{hegselmann2023tabllm}
Stefan Hegselmann, Alejandro Buendia, Hunter Lang, Monica Agrawal, Xiaoyi
  Jiang, and David Sontag.
\newblock Tabllm: Few-shot classification of tabular data with large language
  models, 2023.

\bibitem[Hendrycks et~al.(2021)Hendrycks, Burns, Kadavath, Arora, Basart, Tang,
  Song, and Steinhardt]{gpt-maths}
Dan Hendrycks, Collin Burns, Saurav Kadavath, Akul Arora, Steven Basart, Eric
  Tang, Dawn Song, and Jacob Steinhardt.
\newblock Measuring mathematical problem solving with the {MATH} dataset.
\newblock \emph{CoRR}, abs/2103.03874, 2021.
\newblock URL \url{https://arxiv.org/abs/2103.03874}.

\bibitem[Hollmann et~al.(2022)Hollmann, M{\"u}ller, Eggensperger, and
  Hutter]{hollmann2022tabpfn}
Noah Hollmann, Samuel M{\"u}ller, Katharina Eggensperger, and Frank Hutter.
\newblock Tabpfn: A transformer that solves small tabular classification
  problems in a second.
\newblock \emph{arXiv preprint arXiv:2207.01848}, 2022.

\bibitem[Horn et~al.(2019)Horn, Pack, and Rieger]{horn2019autofeat}
Franziska Horn, Robert Pack, and Michael Rieger.
\newblock The autofeat python library for automatic feature engineering and
  selection.
\newblock \emph{CoRR}, abs/1901.07329, 2019.
\newblock URL \url{http://arxiv.org/abs/1901.07329}.

\bibitem[Hutter et~al.(2019)Hutter, Kotthoff, and Vanschoren]{hutter-book19a}
F.~Hutter, L.~Kotthoff, and J.~Vanschoren (eds.).
\newblock \emph{Automated Machine Learning: Methods, Systems, Challenges}.
\newblock Springer, 2019.
\newblock Available for free at http://automl.org/book.

\bibitem[Kanter \& Veeramachaneni(2015)Kanter and
  Veeramachaneni]{kanter2015deep}
James~Max Kanter and Kalyan Veeramachaneni.
\newblock Deep feature synthesis: Towards automating data science endeavors.
\newblock In \emph{2015 IEEE international conference on data science and
  advanced analytics (DSAA)}, pp.\  1--10. IEEE, 2015.

\bibitem[Khurana et~al.(2016)Khurana, Turaga, Samulowitz, and
  Parthasrathy]{khurana2016cognito}
Udayan Khurana, Deepak Turaga, Horst Samulowitz, and Srinivasan Parthasrathy.
\newblock Cognito: Automated feature engineering for supervised learning.
\newblock In \emph{2016 IEEE 16th International Conference on Data Mining
  Workshops (ICDMW)}, pp.\  1304--1307, 2016.
\newblock \doi{10.1109/ICDMW.2016.0190}.

\bibitem[Khurana et~al.(2018)Khurana, Samulowitz, and
  Turaga]{khurana2018feature}
Udayan Khurana, Horst Samulowitz, and Deepak Turaga.
\newblock Feature engineering for predictive modeling using reinforcement
  learning.
\newblock In \emph{Proceedings of the AAAI Conference on Artificial
  Intelligence}, volume~32, 2018.

\bibitem[Lee \& Macke(2020)Lee and Macke]{lee2020human}
Doris Jung-Lin Lee and Stephen Macke.
\newblock A human-in-the-loop perspective on automl: Milestones and the road
  ahead.
\newblock \emph{IEEE Data Engineering Bulletin}, 2020.

\bibitem[Narayan et~al.(2022)Narayan, Chami, Orr, Arora, and
  Ré]{narayan2022foundation}
Avanika Narayan, Ines Chami, Laurel Orr, Simran Arora, and Christopher Ré.
\newblock Can foundation models wrangle your data?, 2022.

\bibitem[Nargesian et~al.(2017)Nargesian, Samulowitz, Khurana, Khalil, and
  Turaga]{nargesian2017learning}
Fatemeh Nargesian, Horst Samulowitz, Udayan Khurana, Elias~B Khalil, and
  Deepak~S Turaga.
\newblock Learning feature engineering for classification.
\newblock In \emph{Ijcai}, volume~17, pp.\  2529--2535, 2017.

\bibitem[OpenAI(2023{\natexlab{a}})]{openai2023gpt4}
OpenAI.
\newblock Gpt-4 technical report, 2023{\natexlab{a}}.

\bibitem[OpenAI(2023{\natexlab{b}})]{openaiop3:online}
OpenAI.
\newblock openai/openai-cookbook: Examples and guides for using the openai api.
\newblock \url{https://github.com/openai/openai-cookbook}, 2023{\natexlab{b}}.
\newblock (Accessed on 04/20/2023).

\bibitem[{OpenAI Community}(2021)]{OpenAIHaiku}
{OpenAI Community}.
\newblock {GPT-3 can’t count syllables - or doesn’t “get” haiku}.
\newblock
  \url{https://community.openai.com/t/gpt-3-cant-count-syllables-or-doesnt-get-haiku/18733},
  2021.
\newblock Accessed on: 2023-03-21.

\bibitem[Prabhu \& Birhane(2020)Prabhu and Birhane]{prabhu2020large}
Vinay~Uday Prabhu and Abeba Birhane.
\newblock Large image datasets: A pyrrhic win for computer vision?, 2020.

\bibitem[Raji et~al.(2021)Raji, Bender, Paullada, Denton, and
  Hanna]{raji2021ai}
Inioluwa~Deborah Raji, Emily~M. Bender, Amandalynne Paullada, Emily Denton, and
  Alex Hanna.
\newblock Ai and the everything in the whole wide world benchmark, 2021.

\bibitem[Rohlf(2023)]{AI_Code_Generation_and_Cybersecurity_2023}
Chris Rohlf.
\newblock Ai code generation and cybersecurity, April 2023.
\newblock URL
  \url{https://www.cfr.org/blog/ai-code-generation-and-cybersecurity}.

\bibitem[Talboy \& Fuller(2023)Talboy and Fuller]{talboy2023challenging}
Alaina~N. Talboy and Elizabeth Fuller.
\newblock Challenging the appearance of machine intelligence: Cognitive bias in
  llms and best practices for adoption, 2023.

\bibitem[Vanschoren et~al.(2013)Vanschoren, van Rijn, Bischl, and
  Torgo]{OpenML2013}
Joaquin Vanschoren, Jan~N. van Rijn, Bernd Bischl, and Luis Torgo.
\newblock Openml: Networked science in machine learning.
\newblock \emph{SIGKDD Explorations}, 15\penalty0 (2):\penalty0 49--60, 2013.
\newblock \doi{10.1145/2641190.2641198}.
\newblock URL \url{http://doi.acm.org/10.1145/2641190.2641198}.

\bibitem[Vanschoren et~al.(2014)Vanschoren, van Rijn, Bischl, and
  Torgo]{VanschorenRBT14}
Joaquin Vanschoren, Jan~N. van Rijn, Bernd Bischl, and Lu{\'{\i}}s Torgo.
\newblock Openml: networked science in machine learning.
\newblock \emph{CoRR}, abs/1407.7722, 2014.
\newblock URL \url{http://arxiv.org/abs/1407.7722}.

\bibitem[Vaswani et~al.(2017)Vaswani, Shazeer, Parmar, Uszkoreit, Jones, Gomez,
  Kaiser, and Polosukhin]{VaswaniSPUJGKP17}
Ashish Vaswani, Noam Shazeer, Niki Parmar, Jakob Uszkoreit, Llion Jones,
  Aidan~N. Gomez, Lukasz Kaiser, and Illia Polosukhin.
\newblock Attention is all you need.
\newblock \emph{CoRR}, abs/1706.03762, 2017.
\newblock URL \url{http://arxiv.org/abs/1706.03762}.

\bibitem[Vos et~al.(2022)Vos, D{\"o}hmen, and Schelter]{vos2022towards}
David Vos, Till D{\"o}hmen, and Sebastian Schelter.
\newblock Towards parameter-efficient automation of data wrangling tasks with
  prefix-tuning.
\newblock In \emph{NeurIPS 2022 First Table Representation Workshop}, 2022.
\newblock URL \url{https://openreview.net/forum?id=8kyYJs2YkFH}.

\bibitem[Weerts et~al.(2023)Weerts, Pfisterer, Feurer, Eggensperger, Bergman,
  Awad, Vanschoren, Pechenizkiy, Bischl, and Hutter]{weerts2023fairness}
Hilde Weerts, Florian Pfisterer, Matthias Feurer, Katharina Eggensperger,
  Edward Bergman, Noor Awad, Joaquin Vanschoren, Mykola Pechenizkiy, Bernd
  Bischl, and Frank Hutter.
\newblock Can fairness be automated? guidelines and opportunities for
  fairness-aware automl, 2023.

\bibitem[Wei et~al.(2021)Wei, Bosma, Zhao, Guu, Yu, Lester, Du, Dai, and
  Le]{wei2021finetuned}
Jason Wei, Maarten Bosma, Vincent~Y Zhao, Kelvin Guu, Adams~Wei Yu, Brian
  Lester, Nan Du, Andrew~M Dai, and Quoc~V Le.
\newblock Finetuned language models are zero-shot learners.
\newblock \emph{arXiv preprint arXiv:2109.01652}, 2021.

\bibitem[Wei et~al.(2023)Wei, Wang, Schuurmans, Bosma, Ichter, Xia, Chi, Le,
  and Zhou]{wei2023chainofthought}
Jason Wei, Xuezhi Wang, Dale Schuurmans, Maarten Bosma, Brian Ichter, Fei Xia,
  Ed~Chi, Quoc Le, and Denny Zhou.
\newblock Chain-of-thought prompting elicits reasoning in large language
  models, 2023.

\bibitem[{W}es {M}c{K}inney(2010)]{mckinney-proc-scipy-2010}
{W}es {M}c{K}inney.
\newblock {D}ata {S}tructures for {S}tatistical {C}omputing in {P}ython.
\newblock In {S}t\'efan van~der {W}alt and {J}arrod {M}illman (eds.),
  \emph{{P}roceedings of the 9th {P}ython in {S}cience {C}onference}, pp.\  56
  -- 61, 2010.
\newblock \doi{10.25080/Majora-92bf1922-00a}.

\bibitem[Wold et~al.(1987)Wold, Esbensen, and Geladi]{wold1987principal}
Svante Wold, Kim Esbensen, and Paul Geladi.
\newblock Principal component analysis.
\newblock \emph{Chemometrics and intelligent laboratory systems}, 2\penalty0
  (1-3):\penalty0 37--52, 1987.

\bibitem[Zhang et~al.(2019)Zhang, Hao, Fogelman-Souli{\'e}, and
  Wang]{zhang2019automatic}
Jianyu Zhang, Jianye Hao, Fran{\c{c}}oise Fogelman-Souli{\'e}, and Zan Wang.
\newblock Automatic feature engineering by deep reinforcement learning.
\newblock In \emph{Proceedings of the 18th International Conference on
  Autonomous Agents and MultiAgent Systems}, pp.\  2312--2314, 2019.

\bibitem[Ziegler et~al.(2019)Ziegler, Stiennon, Wu, Brown, Radford, Amodei,
  Christiano, and Irving]{ziegler2019fine}
Daniel~M Ziegler, Nisan Stiennon, Jeffrey Wu, Tom~B Brown, Alec Radford, Dario
  Amodei, Paul Christiano, and Geoffrey Irving.
\newblock Fine-tuning language models from human preferences.
\newblock \emph{arXiv preprint arXiv:1909.08593}, 2019.

\end{thebibliography}
\bibliographystyle{iclr2023_conference_tinypaper}

\newpage
\appendix

\section{Acknowledgements}
GPT-4 \citep{openai2023gpt4} was used in the following ways: to help us iterate on LaTeX formatting and diagram plotting; for text summarization; and as a copyediting tool; for rephrasing proposals.

\section{Broader Impact Statement}
\label{sec:impacts}

\subsection{Social Impact of Automation} The broader implications of our research may contribute to the automation of data science tasks, potentially displacing workers in the field. However, \methodname{} crucially depends on the users inputs for feature generation and processing and provides an example of human-in-the-loop AutoML. The automation of routine tasks could free up data scientists to focus on higher-level problem-solving and decision-making activities. It is essential for stakeholders to be aware of these potential consequences, and to consider strategies for workforce education and adaptation to ensure a smooth transition as AI technologies continue to evolve.

\subsection{Replication of Biases} AI algorithms have been observed to replicate and perpetuate biases observed in their training data distribution. \methodname{} leverages GPT-4, which has been trained on web crawled data that contains existing social biases and generated features may be biased on these biases. An example study of such biases in web-crawled data was done by \citet{prabhu2020large}. When data that contains demographic information or other data that can potentially be used to discriminate against groups, we advise not to use \methodname{} or to proceed with great caution, double checking the generated features.

We would like to mention the best practice studies on fairness for AutoML by \citet{weerts2023fairness} and on bias in the use of LLMs by \citet{talboy2023challenging}.

\citet{weerts2023fairness} give recommendations for what AutoML developers can do to improve fairness-conscious AutoML, and we follow the three most emphasized closely.
i) We clearly emphasize the limitations and biases of our method.
i) We emphasize the `Usefulness' comment that our prompt asks the LLM to write into our UI to ``help users identify sources of fairness-related harm'', as \citet{weerts2023fairness} write.
ii) We execute code only after asking the user for explicit confirmation that it does not contain dangerous actions or biases; these kinds of design decisions are called "seam-full design" by \citep{weerts2023fairness}.

\citep{talboy2023challenging} give recommendations on how to best use LLMs. Their recommendation, which we as interface developers can more easily follow, is the one they emphasize the most. Namely, that ``LLMs [...] should be used as decision support tools, not final decision makers.
We follow this closely by not allowing the LLM to execute code without the user's explicit consent.

The choice of benchmarks can also introduce dangerous biases in machine learning methods, as there are only a few that dominate the landscape \citep{raji2021ai}. We address this problem by using datasets from two different sources (Kaggle and OpenML). In addition, we use multiple datasets from each source.

\paragraph{How Can Biases Arise in CAAFE?}
Since we are using a trained model (the LLM) to define the training setup (features used) of a downstream classifier trained on potentially biased data, we can have bias at three levels:
i) at the level of the LLM, and
ii) at the level of the generated features,
ii) at the level of the downstream classifier.

The level of the downstream classifier is affected by the previous two levels, but is not directly affected by CAAFE, so we do not discuss it in detail. Here, as in any data science application, the user must be careful to use unbiased data.
The way we set up CAAFE has some robustness against biases in the first two steps, which are controlled by CAAFE:

i) We use GPT-4, where great care has been taken to mitigate biases – still it does contain biases in its generations. We add a statement that choosing a language model that is trained to mitigate biases is a crucial step for the user to tackle this problem. 

ii) Feature engineering operations by the LLM are only retained if they lead to improvements in cross-validation, due to the way our method is set up (we generate code and verify that it improves performance at each step).
This gives us a slightly better defense against bias than using the LLM outputs directly, since biases that are present in the LLM but not in the data are discarded.

However, since the biases of the LLM and the dataset are most likely not independent, there is still a risk that the two will be aligned.
To illustrate this risk and make users aware of potential problems, we include an illustrative Example (see end of this reply).

\paragraph{Simple Example}
We have built a fake dataset that has only one feature, a person's name, and as output we want to predict whether the person is a doctor or a nurse.
The description of the dataset used in our challenge is

\begin{lstlisting}
Doctor-or-Nurse is a dataset that asks to predict whether a person is a doctor or a nurse just based on their name.
```
The intentionally biased sample of the dataset shown to the LLM in the prompt is
```
Columns in `df` (true feature dtypes listed here, categoricals encoded as int):
Name (object): NaN-freq [0.0%], Samples ['Anna', 'Jack', 'Frank', 'Laura', 'Travis', 'Sophia']
is_doctor (int): NaN-freq [0.0%], Samples [0, 1, 1, 0, 1, 0]
\end{lstlisting}
In all cases in this sample a name commonly associated with male gender is associated with a doctor.
Additionally, we only used female names ending in an "a".

We can see that GPT-4 tends to suggest an attribute that uses the typically gender-associated ending of the name to classify, i.e. it generates the following output

\begin{lstlisting}[language=Python, caption=CAAFE output]
# Feature: end_with_a
# Usefulness: Names ending with "a" might be more common in females. Assuming there might be a gender bias in the doctor or nurse profession, this could provide some useful information.
# Input samples: ['Anna', 'Jack', 'Frank']
df['end_with_a'] = df['Name'].str.endswith('a').astype(int)
\end{lstlisting}

We can see that our prompt actually makes it very easy to detect the bias of the LLM in this case, since the `Usefulness` comment the LLM is asked to provide, at least in this non-cherry-picked example, just gives away its bias.

\subsection{AI Model Interpretability} As the adoption of advanced AI methods grows, it becomes increasingly important to comprehend and interpret their results. Our approach aims to enhance interpretability by providing clear explanations of model outputs and generating simple code, thus making the automated feature engineering process more transparent.

\subsection{Risk of increasing AI capabilities} We do not believe this research affects the general capabilities of LLMs but rather demonstrates their application. As such we estimate our work does not contribute to the risk of increasing AI capabilities.

\section{Reproducibility}\label{sec:reproducibility}

\paragraph{Code release} In an effort to ensure reproducibility, we release code to reproduce our experiments at %\url{https://github.com/cafeautomatedfeatures/CAFE}
\url{https://github.com/automl/CAAFE}
We release a minimal demo at \href{https://colab.research.google.com/drive/1JPgiEAKO1e7aofm2vqXA1yTHRhAqZgTx}{a simple demo}.

\textbf{Availability of datasets} All datasets used in our experiments are freely available at \url{OpenML.org}~\citep{VanschorenRBT14} or at \url{kaggle.com}, with downloading procedures included in the submission.

\section{Full LLM Prompt}

Figure \ref{fig:llm_prompt} shows the full prompt for one examplary dataset. The generated prompts are in our repository: \url{https://github.com/automl/CAAFE/tree/main/data/generated_code}.
% Anonymized: https://github.com/cafeautomatedfeatures/CAFE/tree/main/data/generated_code
\begin{figure}[H]
    \centering
    \begin{minipage}{\textwidth}
        \input{figures/llm_prompt}
    \end{minipage}
    \caption{Full LLM Prompt for the CMC dataset. The generated code will be the reply to this prompt.}
    \label{fig:llm_prompt}
\end{figure}

\section{Additional Results}

\subsection{Semantic Blinding}
Semantic information, i.e. the context of the dataset and its columns, is crucial and can only be captured through laborious human work or our novel approach of using LLMs - this is the core of our approach. To further verify and quantify this claim, we perform an experiment where the context of the dataset is left out (i.e. feature names and dataset description are not given to the LLM). We find a strong drop in performance from an average AUROC of 0.822 to 0.8 over all datasets for GPT-4.

\begin{table}[H]
    \centering
    \caption{CAAFE with semantic and without "Semantic Blinding". For "Semantic Blinding" feature names and the dataset description is concealed to CAAFE. Mean ROC AUC and average rank (ROC AUC) per downstream classification method and feature extension method is shown. Best performing AutoFE method per classifier is shown in bild. The features generated by \methodname{} are chosen with TabPFN as classifier. Ranks are calculated across all classifiers and feature engineering methods.}
    \scriptsize
    \begin{table}[H]
    \label{table:proposal}
    \centering
    \scriptsize
    \setlength{\tabcolsep}{5pt}
\begin{tabular}{ll|l|ll|ll}
\toprule
{} & {} & {} & \multicolumn{2}{c}{GPT-3.5} & \multicolumn{2}{c}{GPT-4} \\
{} & {} & {No FE} & Semantic Blinding & Default & Semantic Blinding & Default \\

\midrule
Log. Reg. & Mean &  0.749 &      0.754 &          0.763 &      0.749 &  \textbf{0.769} \\
\rowcolor{gray!25}            & Mean Rank &   19.6 &       19.1 &           17.8 &       19.3 &   \textbf{17.4} \\
\midrule
Random Forest & Mean &  0.782 &      0.789 &           0.79 &      0.783 &  \textbf{0.803} \\
\rowcolor{gray!25}            & Mean Rank &   17.2 &       16.5 &           16.8 &       16.5 &   \textbf{14.6} \\
\midrule
ASKL2 & Mean &  0.807 &      0.812 &          0.815 &      0.809 &  \textbf{0.818} \\
\rowcolor{gray!25}            & Mean Rank &   9.17 &       9.29 &  \textbf{8.16} &       8.88 &            8.59 \\
\midrule
Autogluon & Mean &  0.796 &      0.803 &          0.803 &      0.801 &  \textbf{0.812} \\
\rowcolor{gray!25}            & Mean Rank &   12.9 &       12.2 &           11.6 &       12.7 &   \textbf{10.3} \\
\midrule
TabPFN & Mean &  0.798 &      0.807 &          0.806 &        0.8 &  \textbf{0.822} \\
\rowcolor{gray!25}            & Mean Rank &   10.5 &       9.33 &           9.71 &       9.33 &   \textbf{7.59} \\
\bottomrule
\end{tabular}

\end{table}
\end{table}

\subsection{Per Dataset Results}
\label{sec:per_dataset_results}
\begin{table}[H]
    \centering
    \scriptsize
    \caption{ROC AUC OVO results per dataset and downstream classification method. \methodname{} optimized for strong performance on TabPFN.}
    \label{table:results_table_per_dataset_plus_baselines}
    \begin{tabular}{ll|ll|ll|ll}
\toprule
{} & \multicolumn{1}{c}{} & \multicolumn{2}{c}{AutoFE-Base} & \multicolumn{2}{c}{CAAFE} & \multicolumn{2}{c}{AutoFE-Base + CAAFE} \\
{} &  & AutoFeat & DFS & GPT-3.5 & GPT-4 & GPT-4 + AF & GPT-4 + DFS \\
\midrule
airlines                   &  \textbf{0.6211} {\tiny $\pm$.04} &           0.6076 {\tiny $\pm$.04} &   0.595 {\tiny $\pm$.04} &            0.619 {\tiny $\pm$.04} &           0.6203 {\tiny $\pm$.04} &            0.602 {\tiny $\pm$.04} &           0.5966 {\tiny $\pm$.04} \\
balance-scale              &           0.8444 {\tiny $\pm$.29} &           0.8438 {\tiny $\pm$.30} &  0.8428 {\tiny $\pm$.31} &            0.844 {\tiny $\pm$.31} &   \textbf{0.882} {\tiny $\pm$.26} &           0.8812 {\tiny $\pm$.27} &           0.8773 {\tiny $\pm$.27} \\
breast-w                   &           0.9783 {\tiny $\pm$.02} &           0.9713 {\tiny $\pm$.03} &  0.9783 {\tiny $\pm$.02} &  \textbf{0.9809} {\tiny $\pm$.02} &  \textbf{0.9809} {\tiny $\pm$.02} &           0.9713 {\tiny $\pm$.03} &  \textbf{0.9809} {\tiny $\pm$.02} \\
cmc                        &           0.7375 {\tiny $\pm$.02} &           0.7384 {\tiny $\pm$.02} &  0.7349 {\tiny $\pm$.02} &           0.7383 {\tiny $\pm$.02} &  \textbf{0.7393} {\tiny $\pm$.02} &           0.7386 {\tiny $\pm$.02} &           0.7362 {\tiny $\pm$.02} \\
credit-g                   &           0.7824 {\tiny $\pm$.03} &           0.7819 {\tiny $\pm$.03} &  0.7824 {\tiny $\pm$.03} &           0.7824 {\tiny $\pm$.03} &           0.7832 {\tiny $\pm$.03} &   \textbf{0.784} {\tiny $\pm$.03} &           0.7824 {\tiny $\pm$.03} \\
diabetes                   &           0.8427 {\tiny $\pm$.03} &           0.8414 {\tiny $\pm$.03} &  0.8417 {\tiny $\pm$.03} &  \textbf{0.8434} {\tiny $\pm$.03} &           0.8425 {\tiny $\pm$.03} &           0.8432 {\tiny $\pm$.03} &           0.8382 {\tiny $\pm$.03} \\
eucalyptus                 &           0.9319 {\tiny $\pm$.01} &           0.9321 {\tiny $\pm$.01} &  0.9319 {\tiny $\pm$.01} &           0.9317 {\tiny $\pm$.01} &           0.9319 {\tiny $\pm$.00} &  \textbf{0.9323} {\tiny $\pm$.01} &           0.9319 {\tiny $\pm$.01} \\
jungle\_chess..             &           0.9334 {\tiny $\pm$.01} &           0.9197 {\tiny $\pm$.01} &  0.9284 {\tiny $\pm$.01} &           0.9361 {\tiny $\pm$.01} &           0.9453 {\tiny $\pm$.01} &  \textbf{0.9535} {\tiny $\pm$.01} &             0.94 {\tiny $\pm$.01} \\
$\langle Kaggle\rangle$ health-insurance  &           0.5745 {\tiny $\pm$.02} &  \textbf{0.5805} {\tiny $\pm$.03} &  0.5753 {\tiny $\pm$.02} &           0.5745 {\tiny $\pm$.02} &           0.5748 {\tiny $\pm$.02} &           0.5777 {\tiny $\pm$.03} &           0.5782 {\tiny $\pm$.03} \\
$\langle Kaggle\rangle$ pharyngitis       &           0.6976 {\tiny $\pm$.03} &           0.6976 {\tiny $\pm$.03} &  0.6976 {\tiny $\pm$.03} &           0.6976 {\tiny $\pm$.03} &  \textbf{0.7078} {\tiny $\pm$.04} &           0.7073 {\tiny $\pm$.04} &           0.6976 {\tiny $\pm$.03} \\
$\langle Kaggle\rangle$ kidney-stone      &           0.7883 {\tiny $\pm$.04} &           0.7856 {\tiny $\pm$.04} &  0.7929 {\tiny $\pm$.04} &           0.7873 {\tiny $\pm$.04} &           0.7903 {\tiny $\pm$.04} &           0.7875 {\tiny $\pm$.04} &  \textbf{0.7967} {\tiny $\pm$.03} \\
$\langle Kaggle\rangle$ spaceship-titanic &            0.838 {\tiny $\pm$.02} &           0.8486 {\tiny $\pm$.02} &  0.8443 {\tiny $\pm$.02} &           0.8383 {\tiny $\pm$.02} &           0.8405 {\tiny $\pm$.02} &   \textbf{0.853} {\tiny $\pm$.02} &           0.8486 {\tiny $\pm$.02} \\
pc1                        &           0.9035 {\tiny $\pm$.01} &           0.9046 {\tiny $\pm$.01} &  0.9035 {\tiny $\pm$.01} &           0.9087 {\tiny $\pm$.02} &  \textbf{0.9093} {\tiny $\pm$.01} &            0.908 {\tiny $\pm$.01} &           0.9035 {\tiny $\pm$.01} \\
tic-tac-toe                &           0.6989 {\tiny $\pm$.08} &           0.6989 {\tiny $\pm$.08} &  0.6291 {\tiny $\pm$.10} &           0.6989 {\tiny $\pm$.08} &  \textbf{0.9536} {\tiny $\pm$.06} &  \textbf{0.9536} {\tiny $\pm$.06} &            0.938 {\tiny $\pm$.06} \\
\bottomrule
\end{tabular}
\end{table}

\subsection{Generated Prompts and Code}
\label{ssec:prompts_and_code}
You can find the generated prompts and the respective LLM generated code in our repository: \url{
https://github.com/cafeautomatedfeatures/CAFE/tree/main/data/generated_code}.

\section{Compute}
\label{sec:results_cost_and_time}
\begin{figure}[H]
    \centering
    \small
    \includegraphics[scale=0.7]{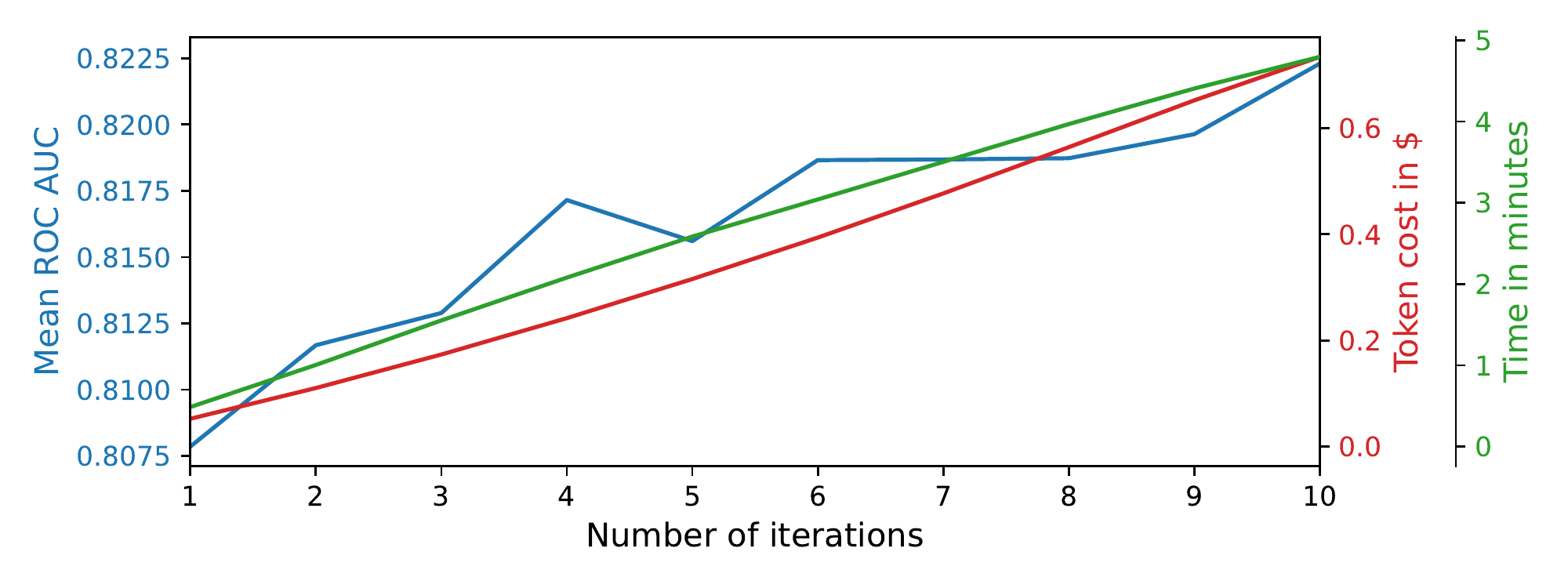}
    \label{fig:results_cost_and_time_fig}
    \caption{Mean ROC AUC OVO, inference cost for GPT and time spent with an increasing number of feature generation runs.}
\end{figure}
Figure \ref{fig:results_cost_and_time_fig} illustrates the increasing performance but also cost and time spent for more feature engineering iterations. Prediction for LLMs is done per token and so the generation of code takes dominates the 4:43 minutes evaluation time of \methodname{} on average per dataset. For GPT-3.5 this time is reduce to about 1/4. Also for GPT-3.5 the cost is reduced to 1/10 as of the writing of this paper. For the evaluation of TabPFN we use one Nvidia RTX 2080 Ti as well as 8 Intel(R) Xeon(R) Gold 6242 CPU @ 2.80GHz CPU cores.

\section{Datasets}\label{app:datasets}
\begin{table}[H]
    \centering
    \small
    \begin{tabular}{lrrrl}
\toprule
{} &  \# Features &  \# Samples &  \# Classes & OpenML ID / Kaggle Name \\
Name                                             &             &            &            &                         \\
\midrule
balance-scale                                    &           4 &        125 &          3 &                      11 \\
breast-w                                         &           9 &         69 &          2 &                      15 \\
cmc                                              &           9 &       1473 &          3 &                      23 \\
credit-g                                         &          20 &       1000 &          2 &                      31 \\
diabetes                                         &           8 &        768 &          2 &                      37 \\
tic-tac-toe                                      &           9 &         95 &          2 &                      50 \\
eucalyptus                                       &          19 &        736 &          5 &                     188 \\
% wine                                             &          13 &         17 &          2 &                     973 \\
pc1                                              &          21 &       1109 &          2 &                    1068 \\
airlines                                         &           7 &       2000 &          2 &                    1169 \\
jungle\_chess\_2pcs\_raw\_endgame\_complete \hspace{-1cm}           &           6 &       2000 &          3 &                   41027 \\
pharyngitis                               &          19 &        512 &          2 &                       \href{https://www.kaggle.com/datasets/yoshifumimiya/pharyngitis}{\textit{pharyngitis}}  \\
health-insurance &          13 &       2000 &          2 &          \href{https://www.kaggle.com/datasets/owaiskhan9654/health-insurance-lead-prediction-raw-data}{\textit{health-insurance-lead-prediction-raw-data}}               \\
spaceship-titanic                         &          13 &       2000 &          2 &                       \href{https://www.kaggle.com/competitions/spaceship-titanic}{\textit{spaceship-titanic}}  \\
kidney-stone                   &           7 &        414 &          2 &                        \href{https://www.kaggle.com/competitions/playground-series-s3e12}{\textit{playground-series-s3e12}} \\
\bottomrule
\end{tabular}

    \caption{Test datasets used for the evaluation. See Section \ref{sec:experimental_setup} for a description of the datasets used.   \label{table:test_datasets_table}}
\end{table}

\subsection{Dataset Collection and Preprocessing}
\label{sec:datasets}
\paragraph{OpenML datasets} We use small datasets from OpenML~\citep{OpenML2013,OpenMLPython2019} that have descriptive feature names (i.e. we do not include any datasets with numbered feature names). Datasets on OpenML contain a task description that we provide as user context to our method and that we clean from redundant information for feature engineering, such as author names or release history. While some descriptions are very informative, other descriptions contain much less information. We remove datasets with more than 20 features, since the prompt length rises linearly with the number of features and exceeds the permissible 8,192 tokens that standard GPT-4 can accept. We show all datasets we used in Table \ref{table:test_datasets_table} in Appendix \ref{app:datasets}.
When datasets are perfectly solvable with TabPFN alone (i.e. reaches ROC AUC of 1.0) we reduce the training set size for that dataset to 10\% or 20\% of the original dataset size. This is the case for the datasets ``balance-scale'' (20\%), ``breast-w'' (10\%) and ``tic-tac-toe'' (10\%).
We focus on small datasets with up to $2\,000$ samples in total, because feature engineering is most important and significant for smaller datasets.

\paragraph{Kaggle datasets} 
We additionally evaluate \methodname{} on $4$ datasets from Kaggle that were released after the knowledge cutoff of our LLM Model. These datasets contain string features as well. String features allow for more complex feature transformations, such as separating Names into First and Last Names, which allows grouping families. 
% \todo[inline]{Frank: did you drop rows that contain missing values for our evaluations because they resulted in errors, or why?}
We drop rows that contain missing values for our evaluations. Details of these datasets can also be found in Table \ref{table:test_datasets_table} in Appendix \ref{app:datasets}.

\subsection{Dataset Descriptions}
The dataset descriptions used were crawled from the respective datasource. For OpenML prompts uninformative information such as the source or reference papers were removed. Figures \ref{fig:data_desc} show the parsed dataset descriptions used for each dataset.

\begin{figure}[h]
    \centering
    \begin{minipage}{\textwidth}
    \begin{lstlisting}
**Balance Scale Weight & Distance Database**  
This data set was generated to model psychological experimental results.  Each example is classified as having the balance scale tip to the right, tip to the left, or be balanced. The attributes are the left weight, the left distance, the right weight, and the right distance. The correct way to find the class is the greater of (left-distance * left-weight) and (right-distance * right-weight). If they are equal, it is balanced.

 Attribute description  
The attributes are the left weight, the left distance, the right weight, and the right distance.
\end{lstlisting}
    \end{minipage}
    \caption{Dataset description for balance-scale.}
    \label{fig:data_desc}
\end{figure}
\begin{figure}[h]
    \centering
    \begin{minipage}{\textwidth}
    \begin{lstlisting}
**Breast Cancer Wisconsin (Original) Data Set.** Features are computed from a digitized image of a fine needle aspirate (FNA) of a breast mass. They describe characteristics of the cell nuclei present in the image. The target feature records the prognosis (malignant or benign).
\end{lstlisting}
    \end{minipage}
    \caption{Dataset description for breast-w.}
    \label{fig:data_desc}
\end{figure}
\begin{figure}[h]
    \centering
    \begin{minipage}{\textwidth}
    \begin{lstlisting}
 4. Relevant Information:
    This dataset is a subset of the 1987 National Indonesia Contraceptive
    Prevalence Survey. The samples are married women who were either not 
    pregnant or do not know if they were at the time of interview. The 
    problem is to predict the current contraceptive method choice 
    (no use, long-term methods, or short-term methods) of a woman based 
    on her demographic and socio-economic characteristics.
 
 7. Attribute Information:
 
    1. Wife's age                     (numerical)
    2. Wife's education               (categorical)      1=low, 2, 3, 4=high
    3. Husband's education            (categorical)      1=low, 2, 3, 4=high
    4. Number of children ever born   (numerical)
    5. Wife's religion                (binary)           0=Non-Islam, 1=Islam
    6. Wife's now working?            (binary)           0=Yes, 1=No
    7. Husband's occupation           (categorical)      1, 2, 3, 4
    8. Standard-of-living index       (categorical)      1=low, 2, 3, 4=high
    9. Media exposure                 (binary)           0=Good, 1=Not good
    10. Contraceptive method used     (class attribute)  1=No-use 
                                                         2=Long-term
                                                         3=Short-term
\end{lstlisting}
    \end{minipage}
    \caption{Dataset description for cmc.}
    \label{fig:data_desc}
\end{figure}
\begin{figure}[h]
    \centering
    \begin{minipage}{\textwidth}
    \begin{lstlisting}
**German Credit dataset**  
This dataset classifies people described by a set of attributes as good or bad credit risks.

This dataset comes with a cost matrix: 
``` 
Good  Bad (predicted)  
Good   0    1 (actual)  
Bad    5    0  
```

It is worse to class a customer as good when they are bad (5), than it is to class a customer as bad when they are good (1).  



 Attribute description  

1. Status of existing checking account, in Deutsche Mark.  
2. Duration in months  
3. Credit history (credits taken, paid back duly, delays, critical accounts)  
4. Purpose of the credit (car, television,...)  
5. Credit amount  
6. Status of savings account/bonds, in Deutsche Mark.  
7. Present employment, in number of years.  
8. Installment rate in percentage of disposable income  
9. Personal status (married, single,...) and sex  
10. Other debtors / guarantors  
11. Present residence since X years  
12. Property (e.g. real estate)  
13. Age in years  
14. Other installment plans (banks, stores)  
15. Housing (rent, own,...)  
16. Number of existing credits at this bank  
17. Job  
18. Number of people being liable to provide maintenance for  
19. Telephone (yes,no)  
20. Foreign worker (yes,no)
\end{lstlisting}
    \end{minipage}
    \caption{Dataset description for credit-g.}
    \label{fig:data_desc}
\end{figure}
\begin{figure}[h]
    \centering
    \begin{minipage}{\textwidth}
    \begin{lstlisting}
 4. Relevant Information:
       Several constraints were placed on the selection of these instances from
       a larger database.  In particular, all patients here are females at
       least 21 years old of Pima Indian heritage.  ADAP is an adaptive learning
       routine that generates and executes digital analogs of perceptron-like
       devices.  It is a unique algorithm; see the paper for details.
 
 7. For Each Attribute: (all numeric-valued)
    1. Number of times pregnant
    2. Plasma glucose concentration a 2 hours in an oral glucose tolerance test
    3. Diastolic blood pressure (mm Hg)
    4. Triceps skin fold thickness (mm)
    5. 2-Hour serum insulin (mu U/ml)
    6. Body mass index (weight in kg/(height in m)^2)
    7. Diabetes pedigree function
    8. Age (years)
    9. Class variable (0 or 1)

 Relabeled values in attribute 'class'
    From: 0                       To: tested_negative     
    From: 1                       To: tested_positive
\end{lstlisting}
    \end{minipage}
    \caption{Dataset description for diabetes.}
    \label{fig:data_desc}
\end{figure}
\begin{figure}[h]
    \centering
    \begin{minipage}{\textwidth}
    \begin{lstlisting}
**Tic-Tac-Toe Endgame database**  
This database encodes the complete set of possible board configurations at the end of tic-tac-toe games, where "x" is assumed to have played first.  The target concept is "win for x" (i.e., true when "x" has one of 8 possible ways to create a "three-in-a-row").  
\end{lstlisting}
    \end{minipage}
    \caption{Dataset description for tic-tac-toe.}
    \label{fig:data_desc}
\end{figure}
\begin{figure}[h]
    \centering
    \begin{minipage}{\textwidth}
    \begin{lstlisting}
**Eucalyptus Soil Conservation**  
The objective was to determine which seedlots in a species are best for soil conservation in seasonally dry hill country. Determination is found by measurement of height, diameter by height, survival, and other contributing factors. 
 
It is important to note that eucalypt trial methods changed over time; earlier trials included mostly 15 - 30cm tall seedling grown in peat plots and the later trials have included mostly three replications of eight trees grown. This change may contribute to less significant results.

Experimental data recording procedures which require noting include:
 - instances with no data recorded due to experimental recording procedures
   require that the absence of a species from one replicate at a site was
   treated as a missing value, but if absent from two or more replicates at a
   site the species was excluded from the site's analyses.
 - missing data for survival, vigour, insect resistance, stem form, crown form
   and utility especially for the data recorded at the Morea Station; this 
   could indicate the death of species in these areas or a lack in collection
   of data.  



 Attribute Information  
 
  1.  Abbrev - site abbreviation - enumerated
  2.  Rep - site rep - integer
  3.  Locality - site locality in the North Island - enumerated
  4.  Map_Ref - map location in the North Island - enumerated
  5.  Latitude - latitude approximation - enumerated
  6.  Altitude - altitude approximation - integer
  7.  Rainfall - rainfall (mm pa) - integer
  8.  Frosts - frosts (deg. c) - integer
  9.  Year - year of planting - integer
  10. Sp - species code - enumerated
  11. PMCno - seedlot number - integer
  12. DBH - best diameter base height (cm) - real
  13. Ht - height (m) - real
  14. Surv - survival - integer
  15. Vig - vigour - real
  16. Ins_res - insect resistance - real
  17. Stem_Fm - stem form - real
  18. Crown_Fm - crown form - real
  19. Brnch_Fm - branch form - real
  Class:
  20. Utility - utility rating - enumerated



 Relevant papers

Bulluch B. T., (1992) Eucalyptus Species Selection for Soil Conservation in Seasonally Dry Hill Country - Twelfth Year Assessment  New Zealand Journal of Forestry Science 21(1): 10 - 31 (1991)  

Kirsten Thomson and Robert J. McQueen (1996) Machine Learning Applied to Fourteen Agricultural Datasets. University of Waikato Research Report  
https://www.cs.waikato.ac.nz/ml/publications/1996/Thomson-McQueen-96.pdf + the original publication:
\end{lstlisting}
    \end{minipage}
    \caption{Dataset description for eucalyptus.}
    \label{fig:data_desc}
\end{figure}
\begin{figure}[h]
    \centering
    \begin{minipage}{\textwidth}
    \begin{lstlisting}
Binarized version of the original data set (see version 1). The multi-class target feature is converted to a two-class nominal target feature by re-labeling the majority class as positive ('P') and all others as negative ('N'). Originally converted by Quan Sun.
\end{lstlisting}
    \end{minipage}
    \caption{Dataset description for wine.}
    \label{fig:data_desc}
\end{figure}
\begin{figure}[h]
    \centering
    \begin{minipage}{\textwidth}
    \begin{lstlisting}
**PC1 Software defect prediction**  
One of the NASA Metrics Data Program defect data sets. Data from flight software for earth orbiting satellite. Data comes from McCabe and Halstead features extractors of source code.  These features were defined in the 70s in an attempt to objectively characterize code features that are associated with software quality.



 Attribute Information  

1. loc             : numeric % McCabe's line count of code
2. v(g)            : numeric % McCabe "cyclomatic complexity"
3. ev(g)           : numeric % McCabe "essential complexity"
4. iv(g)           : numeric % McCabe "design complexity"
5. n               : numeric % Halstead total operators + operands
6. v               : numeric % Halstead "volume"
7. l               : numeric % Halstead "program length"
8. d               : numeric % Halstead "difficulty"
9. i               : numeric % Halstead "intelligence"
10. e               : numeric % Halstead "effort"
11. b               : numeric % Halstead 
12. t               : numeric % Halstead's time estimator
13. lOCode          : numeric % Halstead's line count
14. lOComment       : numeric % Halstead's count of lines of comments
15. lOBlank         : numeric % Halstead's count of blank lines
16. lOCodeAndComment: numeric
17. uniq_Op         : numeric % unique operators
18. uniq_Opnd       : numeric % unique operands
19. total_Op        : numeric % total operators
20. total_Opnd      : numeric % total operands
21. branchCount     : numeric % of the flow graph
22. branchCount     : numeric % of the flow graph
23. defects         : {false,true} % module has/has not one or more reported defects



 Relevant papers  

- Shepperd, M. and Qinbao Song and Zhongbin Sun and Mair, C. (2013)
Data Quality: Some Comments on the NASA Software Defect Datasets, IEEE Transactions on Software Engineering, 39.

- Tim Menzies and Justin S. Di Stefano (2004) How Good is Your Blind Spot Sampling Policy? 2004 IEEE Conference on High Assurance
Software Engineering.

- T. Menzies and J. DiStefano and A. Orrego and R. Chapman (2004) Assessing Predictors of Software Defects", Workshop on Predictive Software Models, Chicago
\end{lstlisting}
    \end{minipage}
    \caption{Dataset description for pc1.}
    \label{fig:data_desc}
\end{figure}
\begin{figure}[h]
    \centering
    \begin{minipage}{\textwidth}
    \begin{lstlisting}


Airlines Dataset Inspired in the regression dataset from Elena Ikonomovska. The task is to predict whether a given flight will be delayed, given the information of the scheduled departure.
\end{lstlisting}
    \end{minipage}
    \caption{Dataset description for airlines.}
    \label{fig:data_desc}
\end{figure}
\begin{figure}[h]
    \centering
    \begin{minipage}{\textwidth}
    \begin{lstlisting}
    Description 

This dataset is part of a collection datasets based on the game "Jungle Chess" (a.k.a. Dou Shou Qi). For a description of the rules, please refer to the paper (link attached). The paper also contains a description of various constructed features. As the tablebases are a disjoint set of several tablebases based on which (two) pieces are on the board, we have uploaded all tablebases that have explicit different content:

* Rat vs Rat
* Rat vs Panther
* Rat vs. Lion
* Rat vs. Elephant
* Panther vs. Lion
* Panther vs. Elephant
* Tiger vs. Lion
* Lion vs. Lion
* Lion vs. Elephant
* Elephant vs. Elephant
* Complete (Combination of the above)
* RAW Complete (Combination of the above, containing for both pieces just the rank, file and strength information). This dataset contains a similar classification problem as, e.g., the King and Rook vs. King problem and is suitable for classification tasks. 

(Note that this dataset is one of the above mentioned datasets). Additionally, note that several subproblems are very similar. Having seen a given positions from one of the tablebases arguably gives a lot of information about the outcome of the same position in the other tablebases. 

J. N. van Rijn and J. K. Vis, Endgame Analysis of Dou Shou Qi. ICGA Journal 37:2, 120--124, 2014. ArXiv link: https://arxiv.org/abs/1604.07312
\end{lstlisting}
    \end{minipage}
    \caption{Dataset description for jungle\_chess\_2pcs\_raw\_endgame\_complete.}
    \label{fig:data_desc}
\end{figure}
\begin{figure}[h]
    \centering
    \begin{minipage}{\textwidth}
    \begin{lstlisting}
For the data and objective, it is evident that this is a Binary Classification Problem data in the Tabular Data format.
A policy is recommended to a person when they land on an insurance website, and if the person chooses to fill up a form to apply, it is considered a Positive outcome (Classified as lead). All other conditions are considered Zero outcomes.
\end{lstlisting}
    \end{minipage}
    \caption{Dataset description for Kaggle\_health-insurance-lead-prediction-raw-data.}
    \label{fig:data_desc}
\end{figure}
\begin{figure}[h]
    \centering
    \begin{minipage}{\textwidth}
    \begin{lstlisting}
Group A streptococcus (GAS) infection is a major cause of pediatric pharyngitis, and infection with this organism requires appropriate antimicrobial therapy.

There is controversy as to whether physicians can rely on signs and symptoms to select pediatric patients with pharyngitis who should undergo rapid antigen detection testing (RADT) for GAS .

Our objective was to evaluate the validity of signs and symptoms in the selective testing of children with pharyngitis.

Now, let's use machine learning to analyze whether a diagnosis can be made from the child's symptoms and signs.
Can we predict RADT positive?
\end{lstlisting}
    \end{minipage}
    \caption{Dataset description for Kaggle\_pharyngitis.}
    \label{fig:data_desc}
\end{figure}
\begin{figure}[h]
    \centering
    \begin{minipage}{\textwidth}
    \begin{lstlisting}

Dataset Description
In this competition your task is to predict whether a passenger was transported to an alternate dimension during the Spaceship Titanic's collision with the spacetime anomaly. To help you make these predictions, you're given a set of personal records recovered from the ship's damaged computer system.

File and Data Field Descriptions
train.csv - Personal records for about two-thirds (~8700) of the passengers, to be used as training data.
PassengerId - A unique Id for each passenger. Each Id takes the form gggg_pp where gggg indicates a group the passenger is travelling with and pp is their number within the group. People in a group are often family members, but not always.
HomePlanet - The planet the passenger departed from, typically their planet of permanent residence.
CryoSleep - Indicates whether the passenger elected to be put into suspended animation for the duration of the voyage. Passengers in cryosleep are confined to their cabins.
Cabin - The cabin number where the passenger is staying. Takes the form deck/num/side, where side can be either P for Port or S for Starboard.
Destination - The planet the passenger will be debarking to.
Age - The age of the passenger.
VIP - Whether the passenger has paid for special VIP service during the voyage.
RoomService, FoodCourt, ShoppingMall, Spa, VRDeck - Amount the passenger has billed at each of the Spaceship Titanic's many luxury amenities.
Name - The first and last names of the passenger.
Transported - Whether the passenger was transported to another dimension. This is the target, the column you are trying to predict.
\end{lstlisting}
    \end{minipage}
    \caption{Dataset description for kaggle\_spaceship-titanic.}
    \label{fig:data_desc}
\end{figure}

%\begin{table}[bt]
%    \centering
%    \caption{Meta-Datasets used for developing the prior.}
%    \tiny
%    \input{figures/valid_datasets_table}
%    \label{table:valid_datasets_table}
%\end{table}

\end{document}